\documentclass[10pt,journal,compsoc]{IEEEtran}
\pdfoutput=1
\usepackage{amsmath}
\usepackage{diagbox}
\usepackage{graphics}
\usepackage{subfigure}
\usepackage{epsfig}
\usepackage{threeparttable}
\usepackage{xspace}
\usepackage{siunitx}
\usepackage{graphicx}
\usepackage{amssymb}
\usepackage{threeparttable}
\usepackage{color}
\usepackage[normalem]{ulem}
\usepackage{multirow}
\usepackage{float}
\usepackage{amsfonts}
\usepackage{bm}
\usepackage{array}
\usepackage{colortbl}
\usepackage{diagbox}
\usepackage{rotating}
\usepackage{booktabs}
\usepackage{overpic}
\usepackage{enumitem}
\usepackage{colortbl}
\usepackage{dblfloatfix}

\usepackage[table]{xcolor}
\definecolor{myLightBlue}{RGB}{242, 248, 255}
\definecolor{myLightYellow}{RGB}{249, 245, 240}

\usepackage{longtable}
\usepackage{supertabular}
\usepackage{makecell}

\usepackage{bm}
\usepackage{multirow}
\usepackage[export]{adjustbox}

\usepackage{colortbl}
\definecolor{lightgray}{gray}{.9}
\definecolor{deepgray}{gray}{.8}

\usepackage{threeparttable}
\newcolumntype{I}{!{\vrule width 1pt}}
\makeatletter
\newcommand{\thickhline}{%
    \noalign {\ifnum 0=`}\fi \hrule height 1pt
    \futurelet \reserved@a \@xhline
}
\makeatother
\definecolor{mygray}{gray}{.9}
\usepackage{float}
\usepackage[ruled, vlined]{algorithm2e}
\usepackage{pifont}
\usepackage{ragged2e}

\usepackage[pagebackref=false,breaklinks=true,colorlinks,bookmarks=false]{hyperref}

\definecolor{mygray}{gray}{.9}
\definecolor{mygreen}{RGB}{93,173,85}
\definecolor{mywarning}{RGB}{233,144,61}
\definecolor{DarkRed}{RGB}{0,0,0}
\definecolor{azure}{rgb}{0.0, 0.5, 1.0}
\definecolor{gray}{rgb}{0.3, 0.3, 0.3}
\definecolor{DarkGreen}{RGB}{42,110,63}

\usepackage[capitalize]{cleveref}
\crefname{section}{Sec.}{Secs.}
\crefname{table}{Tab.}{Tabs.}
\crefname{section}{§}{§§}

\makeatletter
\DeclareRobustCommand\onedot{\futurelet\@let@token\@onedot}
\def\@onedot{\ifx\@let@token.\else.\null\fi\xspace}



\begin{document}
\title{3D Human Interaction Generation: A Survey}
\author{Siyuan Fan, Wenke Huang, Xiantao Cai, Bo Du,~\IEEEmembership{Senior Member,~IEEE}

\IEEEcompsocitemizethanks{
\IEEEcompsocthanksitem 
Siyuan Fan, Wenke Huang, Xiantao Cai, Bo Du are with the School of Computer Science, Wuhan University, Wuhan, China. \protect
E-mail:\{fansiyuan, wenkehuang, caixiantao, dubo\}@whu.edu.cn
}
}
\markboth{3D HUMAN INTERACTION GENERATION: A SURVEY}%
{Shell \MakeLowercase{\textit{et al.}}: Bare Demo of IEEEtran.cls for Journals}

\IEEEtitleabstractindextext{
\begin{abstract}
3D human interaction generation has emerged as a key research area, focusing on producing dynamic and contextually relevant interactions between humans and various interactive entities. Recent rapid advancements in 3D model representation methods, motion capture technologies, and generative models have laid a solid foundation for the growing interest in this domain. Existing research in this field can be broadly categorized into three areas: human-scene interaction, human-object interaction, and human-human interaction. Despite the rapid advancements in this area, challenges remain due to the need for naturalness in human motion generation and the accurate interaction between humans and interactive entities. In this survey, we present a comprehensive literature review of human interaction generation, which, to the best of our knowledge, is the first of its kind. We begin by introducing the foundational technologies, including model representations, motion capture methods, and generative models. Subsequently, we introduce the approaches proposed for the three sub-tasks, along with their corresponding datasets and evaluation metrics. Finally, we discuss potential future research directions in this area and conclude the survey. Through this survey, we aim to offer a comprehensive overview of the current advancements in the field, highlight key challenges, and inspire future research works.
\end{abstract}
}

\maketitle
\IEEEdisplaynontitleabstractindextext
\IEEEpeerreviewmaketitle
\newcommand{\digits}{{Digits}}
\newcommand{\mnist}{{MNIST}}
\newcommand{\mnistabbrv}{{M}}
\newcommand{\usps}{{USPS}}
\newcommand{\uspsabbrv}{{U}}
\newcommand{\svhn}{{SVHN}}
\newcommand{\svhnabbrv}{{Sv}}
\newcommand{\syn}{{SYN}}
\newcommand{\synabbrv}{{Sy}}

\newcommand{\officecaltech}{{Office Caltech}}
\newcommand{\caltech}{{Caltech}}
\newcommand{\caltechabbrv}{{Ca}}
\newcommand{\amazon}{{Amazon}}
\newcommand{\amazonabbrv}{{Am}}
\newcommand{\webcam}{{Webcam}}
\newcommand{\webcamabbrv}{{W}}
\newcommand{\dslr}{{DSLR}}
\newcommand{\dslrabbrv}{{D}}

\newcommand{\caltechtfs}{{Caltech-256}}
\newcommand{\officeto}{{Office31}}

\newcommand{\officehome}{{Office-Home}}
\newcommand{\art}{{Art}}
\newcommand{\artabbrv}{{Ar}}
\newcommand{\clipart}{{Clipart}}
\newcommand{\clipartabbrv}{{Cl}}
\newcommand{\product}{{Product}}
\newcommand{\productabbrv}{{P}}
\newcommand{\realworld}{{Real World}}
\newcommand{\realworldabbrv}{{RW}}

\newcommand{\pacs}{{PACS}}
\newcommand{\photo}{{Photo}}
\newcommand{\photoabbrv}{{P}}
\newcommand{\artpainting}{{Art Painting}}
\newcommand{\artpaintingabbrv}{{AP}}
\newcommand{\cartoon}{{Cartoon}}
\newcommand{\cartoonabbrv}{{Ct}}
\newcommand{\sketch}{{Sketch}}
\newcommand{\sketchabbrv}{{Sk}}

\newcommand{\cifar}{{Cifar}}
\newcommand{\weperson}{{WePerson}}
\newcommand{\imagenet}{{ImageNet}}
\newcommand{\cifarhun}{{Cifar-100}}
\newcommand{\cifarten}{{Cifar-10}}
\newcommand{\tyimg}{{Tiny-ImageNet}}
\newcommand{\coco}{{COCO}}
\newcommand{\market}{{Market1501}}
\newcommand{\fashionmnist}{{Fashion-MNIST}}

\newcommand{\simplecnn}{{SimpleCNN}}
\newcommand{\resnet}{{ResNet}}
\newcommand{\resnetten}{{ResNet-10}}
\newcommand{\resnettwelve}{{ResNet-12}}
\newcommand{\resneteighteen}{{ResNet-18}}
\newcommand{\resnettwenty}{{ResNet-20}}
\newcommand{\resnetthirtyfour}{{ResNet-34}}
\newcommand{\resnetfifty}{{ResNet-50}}
\newcommand{\resnetonezeroone}{{ResNet-101}}
\newcommand{\resnetonefivetwo}{{ResNet-152}}

\newcommand{\resnext}{{ResNeXt}}
\newcommand{\ResNextFourEleven}{{ResNeXt4-11}}
\newcommand{\mobilenet}{{MobileNet}}
\newcommand{\efficientnet}{{EfficientNet}}
\newcommand{\densenet}{{DenseNet}}
\newcommand{\convnet}{{Convnet}}
\newcommand{\googlenet}{{GoogLeNet}}
\newcommand{\fbnet}{{Fbnet}}

\newcommand{\sgd}{{SGD}}
\newcommand{\adam}{{Adam}}

\newcommand{\finch}{{Finch}}
\newcommand{\hac}{{HAC}}
\newcommand{\kmeans}{{Kmeans}}
\newcommand{\cka}{{CKA}}
\newcommand{\tsne}{{t-SNE}}
\newcommand{\dbscan}{{DBSCAN}}

\newcommand{\ce}{{CE Loss}}
\newcommand{\kl}{{KL Loss}}
\newcommand{\triplet}{{Triplet Loss}}
\newcommand{\normface}{{Normface}}
\newcommand{\sphereface}{{Sphereface}}
\newcommand{\rce}{{RCE}}
\newcommand{\proxyanchor}{{Proxy-Anchor}}
\newcommand{\atk}{{$AT_k$}}
\newcommand{\centerloss}{{Center}}
\newcommand{\largemargin}{{L-Softmax}}
\newcommand{\arcface}{{ArcFace}}
\newcommand{\polyloss}{{PolyLoss}}

\newcommand{\fedavg}{{FedAvg}}
\newcommand{\fedprox}{{FedProx}}
\newcommand{\fedcurv}{{FedCurv}}
\newcommand{\scaffold}{{SCAFFOLD}}
\newcommand{\feddyn}{FedDyn}
\newcommand{\moon}{{MOON}}
\newcommand{\fedproc}{{FedProc}}
\newcommand{\ccvr}{{CCVR}}
\newcommand{\fedaux}{{FEDAUX}}
\newcommand{\fedmgda}{{FedMGDA+}}
\newcommand{\sparsefed}{{SparseFed}}
\newcommand{\feddc}{{FedDC}}
\newcommand{\pfedme}{{pFedME}}
\newcommand{\dqs}{{DQS}}
\newcommand{\trimmedmedian}{{Trim Median}}
\newcommand{\favor}{{FAVOR}}
\newcommand{\fedproto}{{FedProto}}
\newcommand{\rlr}{{RLR}}
\newcommand{\mccda}{{MCC-DA}}
\newcommand{\provablefl}{{ProvableFL}}
\newcommand{\fedagg}{{FedAgg}}
\newcommand{\feddg}{{FedDG}}
\newcommand{\cerp}{{CerP}}
\newcommand{\ida}{{IDA}}
\newcommand{\dpa}{{DPA}}
\newcommand{\neurotoxin}{{Neurotoxin}}
\newcommand{\ircmsda}{{IRCMSDA}}
\newcommand{\creff}{{CReFF}}
\newcommand{\fedada}{{FedAda}}
\newcommand{\fedsae}{{FedSAE}}
\newcommand{\autofedavg}{{Auto-FedAvg}}
\newcommand{\fedbn}{{FedBN}}
\newcommand{\astraea}{{Astraea}}
\newcommand{\fedacs}{{FedACS}}
\newcommand{\dcadam}{{DC-Adam}}
\newcommand{\aegr}{{AEGR}}
\newcommand{\comda}{{Co-MDA}}
\newcommand{\divfl}{{DivFL}}
\newcommand{\fedss}{{FEDSS}}
\newcommand{\climb}{{CLIMB}}
\newcommand{\fedci}{{FedCI}}
\newcommand{\fccl}{{FCCL}}
\newcommand{\mincost}{{MinCost}}
\newcommand{\fedcg}{{FedCG}}
\newcommand{\dsfl}{{DS-FL}}
\newcommand{\fedlaw}{{FEDLAW}}
\newcommand{\bnpfl}{{BNPFL}}
\newcommand{\fedcams}{{FedCAMS}}
\newcommand{\xormixup}{{XorMixUp}}
\newcommand{\fedspeed}{{FEDSPEED}}
\newcommand{\fada}{{FADA}}
\newcommand{\fedfv}{{FedFV}}
\newcommand{\fedasync}{{FedAsync}}
\newcommand{\fedssl}{{FedSSL}}
\newcommand{\furl}{{FURL}}
\newcommand{\fedmd}{{FedMD}}
\newcommand{\cronus}{{Cronus}}
\newcommand{\fedmdnfdp}{FEDMD-NFDP}
\newcommand{\feddf}{{FedDF}}
\newcommand{\fedrad}{{FedRAD}}
\newcommand{\rhfl}{{RHFL}}
\newcommand{\afl}{{AFL}}
\newcommand{\lgfedavg}{LG-FEDAVG}
\newcommand{\fedrs}{{FedRS}}
\newcommand{\adcol}{{ADCOL}}
\newcommand{\fedmix}{{FedMix}}
\newcommand{\fedgkt}{{FedGKT}}
\newcommand{\flmoe}{{FLMoE}}
\newcommand{\fedopt}{{FedOPT}}
\newcommand{\fml}{{FML}}
\newcommand{\fedufo}{{FedUFO}}
\newcommand{\fedsam}{{FedSAM}}
\newcommand{\knnper}{{kNN-Per}}
\newcommand{\fedlc}{{FedLC}}
\newcommand{\fednova}{{FedNova}}
\newcommand{\pfedla}{{pFedLA}}
\newcommand{\orchestra}{{Orchestra}}
\newcommand{\fedrod}{{FED-ROD}}
\newcommand{\fedsm}{{FedSM}}
\newcommand{\fedmlb}{{FedMLB}}
\newcommand{\fedhenn}{{FedHeNN}}
\newcommand{\fedtwo}{{Fed$^2$}}
\newcommand{\fade}{{FADE}}
\newcommand{\mocha}{{MOCHA}}
\newcommand{\fedpara}{{FedPara}}
\newcommand{\waffle}{{WAFFLe}}
\newcommand{\matcha}{{Matcha}}
\newcommand{\agnosticfl}{{AFL}}
\newcommand{\fedper}{{FEDPER}}
\newcommand{\comt}{{CoMT}}
\newcommand{\fpca}{{FPCA}}
\newcommand{\lotteryfl}{{LotteryFL}}
\newcommand{\rcfl}{{RCFL}}
\newcommand{\fsmafl}{{FSMAFL}}
\newcommand{\fedreid}{{FedReID}}
\newcommand{\fedamp}{{FedAMP}}
\newcommand{\fedheal}{{FedHEAL}}
\newcommand{\pfedhn}{{pFedHN}}
\newcommand{\fedmatch}{{FedMatch}}
\newcommand{\fedbe}{{FEDBE}}
\newcommand{\ditto}{{Ditto}}
\newcommand{\fedrecon}{{FEDRECON}}
\newcommand{\fedgen}{{FEDGEN}}
\newcommand{\fedcgan}{{FedCGAN}}
\newcommand{\fedad}{{FedAD}}
\newcommand{\fedzdac}{{Fed-ZDAC}}
\newcommand{\heterofl}{{HeteroFL}}
\newcommand{\fedfomo}{{FedFomo}}
\newcommand{\soteria}{{Soteria}}
\newcommand{\fedagm}{{FedAGM}}
\newcommand{\fednew}{{FedNew}}
\newcommand{\fedalign}{{FedAlign}}
\newcommand{\feddst}{{FedDST}}
\newcommand{\safe}{{Safe}}
\newcommand{\rrfl}{{RRFL}}
\newcommand{\fedcorr}{{FedCorr}}
\newcommand{\splitmix}{{SplitMix}}
\newcommand{\fedcor}{{FedCor}}
\newcommand{\fedftg}{{FedFTG}}
\newcommand{\fedbabu}{{FedBABU}}
\newcommand{\sfl}{{SFL}}
\newcommand{\gifair}{{GIFAIR}}
\newcommand{\fedsoft}{{FedSoft}}
\newcommand{\fedreg}{{FedReg}}
\newcommand{\fedmat}{{FedMAT}}
\newcommand{\mabrfl}{{MAB-RFL}}
\newcommand{\cfed}{{CFeD}}
\newcommand{\cgpfl}{{CGPFL}}
\newcommand{\maker}{{MaKEr}}
\newcommand{\harmofl}{{HarmoFL}}
\newcommand{\fedpa}{{FEDPA}}
\newcommand{\spahm}{{SPAHM}}
\newcommand{\fedma}{{FedMA}}
\newcommand{\fedntd}{{FedNTD}}
\newcommand{\feddualavg}{{FEDDUALAVG}}
\newcommand{\fedrn}{{FedRN}}
\newcommand{\fedpcl}{{FedPCL}}
\newcommand{\pgfl}{{PGFL}}
\newcommand{\ifca}{IFCA}
\newcommand{\ktpfl}{{KT-pFL}}
\newcommand{\fedscale}{{FedScale}}
\newcommand{\fedfa}{{FedFA}}
\newcommand{\fpl}{{FPL}}
\newcommand{\flhc}{{FL+HC}}
\newcommand{\faug}{{FAug}}
\newcommand{\cfl}{{CFL}}
\newcommand{\mcfl}{{MFCL}}
\newcommand{\multikrum}{{Multi Krum}}
\newcommand{\fedgroup}{{FedGroup}}
\newcommand{\hypcluster}{{HYPCLUSTER}}
\newcommand{\fedpr}{{FedPR}}
\newcommand{\dynafed}{{DYNAFED}}
\newcommand{\fedce}{{FedCE}}
\newcommand{\qffl}{{qFFL}}
\newcommand{\fedfusion}{{FedFusion}}
\newcommand{\dafkd}{{DaFKD}}
\newcommand{\fedpvr}{{FedPVR}}
\newcommand{\feddm}{{FedDM}}
\newcommand{\ot}{{OT}}
\newcommand{\fedsim}{{FedSim}}
\newcommand{\crfl}{{CRFL}}
\newcommand{\smartfl}{{SmartFL}}
\newcommand{\foolsgold}{{FoolsGold}}
\newcommand{\krum}{{Krum}}
\newcommand{\bulyan}{{Bulyan}}
\newcommand{\rfa}{{RFA}}
\newcommand{\rsa}{{RSA}}
\newcommand{\dncagg}{{{DnC}}}
\newcommand{\median}{{Median}}
\newcommand{\afa}{{AFA}}
\newcommand{\aaa}{{AAA}}
\newcommand{\faba}{{FABA}}
\newcommand{\fltrust}{{FLTrust}}
\newcommand{\sageflow}{{Sageflow}}
\newcommand{\dba}{{DBA}}
\newcommand{\rbtm}{{RB-TM}}
\newcommand{\ipm}{{IPM}}
\newcommand{\lie}{{Little Is Enough}}
\newcommand{\creamfl}{{CreamFL}}
\newcommand{\fedseg}{{FedSeg}}
\newcommand{\dimkrum}{{Dim-Krum}}
\newcommand{\fedsi}{{FedSI}}
\newcommand{\jupiter}{{Jupiter}}
\newcommand{\fedtfi}{{FedTFI}}
\newcommand{\fedbeal}{{FedBEAL}}
\newcommand{\fedspace}{{FedSpace}}
\newcommand{\cbafed}{{CBAFed}}
\newcommand{\fedbr}{{FedBR}}
\newcommand{\copa}{{COPA}}
\newcommand{\fedsr}{{FedSR}}
\newcommand{\gradma}{{GradMA}}
\newcommand{\fedga}{{FedGA}}
\newcommand{\fednoro}{{FedNoRo}}
\newcommand{\fedlsr}{{FedLSR}}
\newcommand{\ccst}{{CCST}}
\newcommand{\iopfl}{{IOP-FL}}
\newcommand{\csac}{{CSAC}}
\newcommand{\mckd}{{MCKD}}
\newcommand{\fedinb}{{FedINB}}
\newcommand{\cffl}{{CFFL}}
\newcommand{\fcfl}{{FCFL}}
\newcommand{\fedfaim}{{FedFAIM}}
\newcommand{\cgsv}{{CGSV}}
\newcommand{\apple}{{APPLE}}
\newcommand{\fedthe}{{FedTHE}}
\newcommand{\fedclip}{{FedCLIP}}
\newcommand{\feddecorr}{{FedDecorr}}
\newcommand{\flip}{{FLIP}}
\newcommand{\depthfl}{{DepthFL}}
\newcommand{\fedpac}{{FedPAC}}
\newcommand{\ilrg}{{iLRG}}
\newcommand{\feddar}{{FedDAR}}
\newcommand{\fedhkd}{{FedHKD}}
\newcommand{\batfl}{{BatFL}}
\newcommand{\kdthreea}{{KD3A}}
\newcommand{\fedadg}{{FedADG}}
\newcommand{\fedadgiotj}{{FedADG}}
\newcommand{\fthreeba}{{F3BA}}
\newcommand{\dbfat}{{DBFAT}}
\newcommand{\fedala}{{FedALA}}
\newcommand{\defl}{{DeFL}}
\newcommand{\fedmdfg}{{FedMDFG}}
\newcommand{\fednh}{{FedNH}}
\newcommand{\fedrbn}{{FedBRN}}
\newcommand{\fairfed}{{FairFed}}
\newcommand{\clusterattack}{{ClusterAttack}}
\newcommand{\fang}{{Fang}}

\newcommand{\hfl}{{Horizontal Federated Learning}}
\newcommand{\hflabbrv}{{HFL}}
\newcommand{\vfl}{{Vertical Federated Learning}}
\newcommand{\vflabbrv}{{VFL}}
\newcommand{\ftl}{{Federated Transfer Learning}}
\newcommand{\ftlabbrv}{{FTL}}
\newcommand{\llm}{{Large Language Model}}
\newcommand{\llmabbrv}{{LLM}}

\newcommand{\gfl}{{Generalizable Federated Learning}}
\newcommand{\gflabbrv}{{GFL}}
\newcommand{\crosscal}{{Cross Calibration}}
\newcommand{\crosscalabbrv}{{CorrCal}}
\newcommand{\clientreg}{{Client Regularization}}
\newcommand{\clientregabbrv}{{CliReg}}
\newcommand{\clientaug}{{Client Augmentation}}
\newcommand{\clientaugabbrv}{{CliAug}}
\newcommand{\serverope}{{Server Operation}}
\newcommand{\serveropeabbrv}{{SerOpe}}
\newcommand{\optimcal}{{Client Regularization}}
\newcommand{\optimcalabbrv}{{OptCal}}
\newcommand{\ungene}{{Unknown Generalization}}
\newcommand{\ungeneabbrv}{{UnkGen}}
\newcommand{\fda}{{Federated Domain Adaptation}}
\newcommand{\fdaabbrv}{{FDA}}
\newcommand{\fdg}{{Federated Domain Generalization}}
\newcommand{\fdgabbrv}{{FDG}}
\newcommand{\croclishift}{{Cross-Client Shift}}
\newcommand{\outclishift}{{Out-Client Shift}}

\newcommand{\rfl}{{Robust Federated Learning}}
\newcommand{\rflabbrv}{{RFL}}
\newcommand{\byzantoler}{{Byzantine Tolerance}}
\newcommand{\backdefen}{{Backdoor Defense}}
\newcommand{\distoler}{{Distance Base Tolerance}}
\newcommand{\statoler}{{Statistics Distribution Tolerance}}
\newcommand{\protoler}{{Proxy Dataset Tolerance}}

\newcommand{\refdefen}{{Model Refinement Defense}}
\newcommand{\aggdefen}{{Robust Aggregation Defense}}
\newcommand{\cerdefen}{{Certified Robustness Defense}}

\newcommand{\ffl}{{Fair Federated Learning}}
\newcommand{\fflabbrv}{{FFL}}
\newcommand{\perfair}{{Performance Fairness}}
\newcommand{\perdebioptim}{{Performance Debias Optimization}}
\newcommand{\perfairrewei}{{Performance Debias Reweighting}}
\newcommand{\colfair}{{Collaboration Fairness}}
\newcommand{\indveval}{{Individual Contribution Evaluation}}
\newcommand{\margeval}{{Marginal Contribution Evaluation}}

\newcommand{\poiatt}{{Poisoning Attack}}
\newcommand{\bayatt}{{Byzantine Attack}}
\newcommand{\databayatt}{{Data-Based Byzantine Attack}}
\newcommand{\modelbayatt}{{Model-Based Byzantine Attack}}
\newcommand{\bacatt}{{Backdoor Attack}}
\newcommand{\freerideatt}{{Free Rider Attack}}
\newcommand{\freeridedef}{{Free Rider Detection}}
\newcommand{\outlierparadete}{{Outlier Parameter Detection}}
\newcommand{\contrievaluation}{{Client Contribution Evaluation}}

\newcommand{\priinf}{{Privacy Inference}}
\newcommand{\predbias}{{Prediction Biases}}
\newcommand{\rewaconf}{{Reward Conflict}}

\newcommand{\labelshift}{{Label Skew}}
\newcommand{\domainshift}{{Domain Skew}}

\newcommand{\pairflip}{{Pair Flipping}}
\newcommand{\pairflipabbrv}{{PairF}}
\newcommand{\symflip}{{Symmetry Flipping}}
\newcommand{\symflipabbrv}{{SymF}}
\newcommand{\randomnoise}{{Random Noise}}
\newcommand{\randomnoiseabbrv}{{RanN}}
\newcommand{\addnoise}{{Add Noise}}
\newcommand{\addnoiseabbrv}{{AddN}}
\newcommand{\minmax}{{Min-Max}}
\newcommand{\minmaxabbrv}{{MiMa}}
\newcommand{\minsum}{{Min-Sum}}
\newcommand{\minsumabbrv}{{MiSu}}
\newcommand{\lieabbrv}{{LIE}}

\newcommand{\backdoorabbrv}{{Bac}}
\newcommand{\semanticbackdoorabbrv}{{Sem Bac}}

\newcommand{\croclienacc}{{Cross-Client Accuracy}}
\newcommand{\outclienacc}{{Out-Client Accuracy}}
\newcommand{\accdeclineimp}{{Accuracy Decline Impact}}
\newcommand{\attsuccerat}{{Attack Success Rate}}
\newcommand{\perfovderv}{{Performance Deviation}}
\newcommand{\conmatchdeg}{{Contribution Match Degree}}

\IEEEraisesectionheading{
\section{Introduction}
\label{sec:introduction}
}

\begin{figure}[h!]
	\centering
	\includegraphics[width=0.8\linewidth]{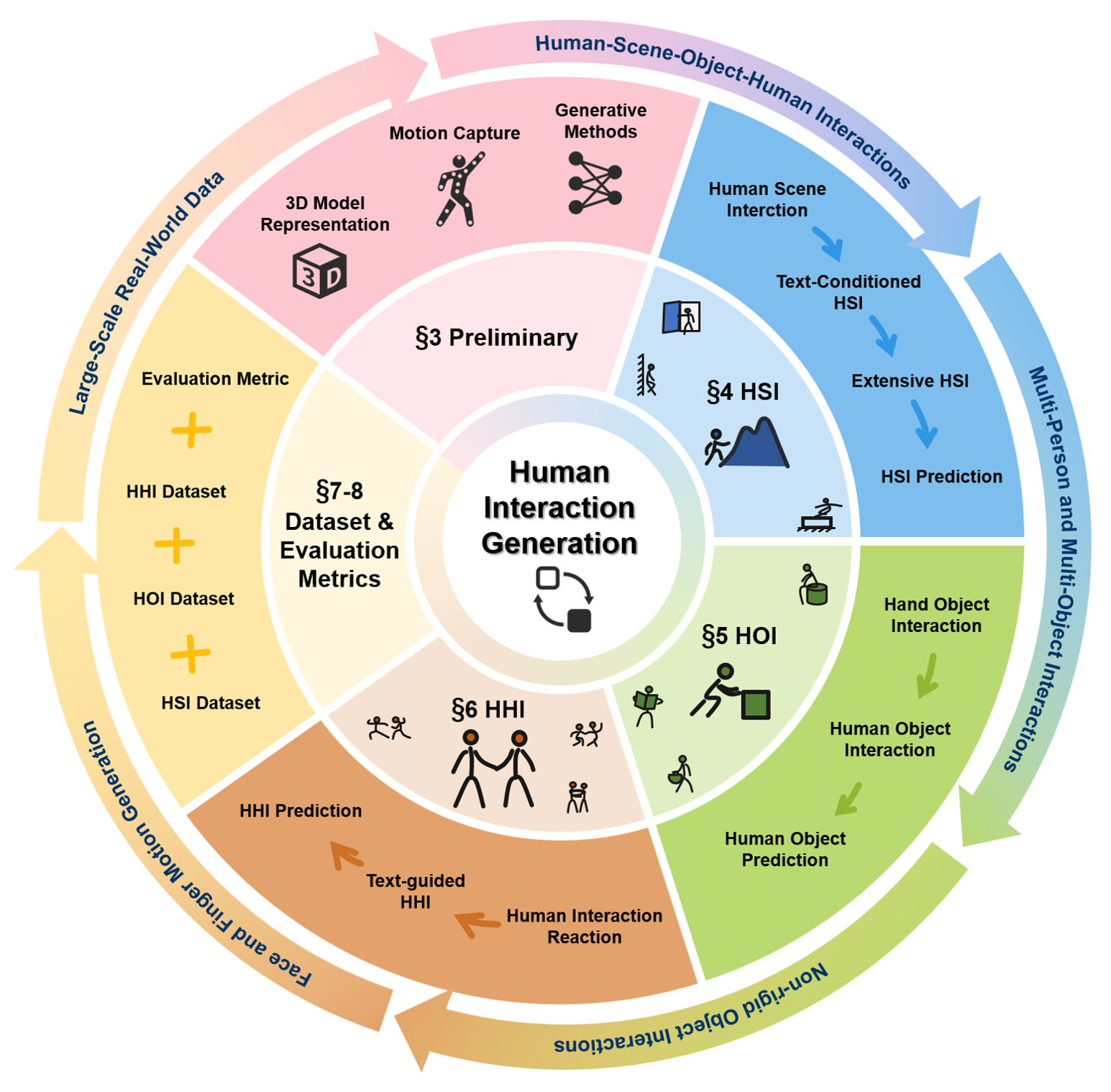}
	\caption{\label{fig:structure}
		This survey begins by defining the relevant techniques for human interaction generation. The core of the survey is organized into three main categories: Human-Scene Interaction (HSI), Human-Object Interaction (HOI), and Human-Human Interaction (HHI). Additionally, this survey summarizes the available datasets for each task and the evaluation metrics.}
\end{figure}

\IEEEPARstart{H}{umans} react to their surrounding environment through interactions, using body language, finger movements, facial expressions, and other non-verbal forms to communicate and influence other objects or individuals in the environment. The goal of human interaction generation is to model and produce both the human motions and the corresponding dynamic movements of the interactive entities involved in the interaction. This technology can be widely applied to VR/AR, video games, animated media, embodied AI, and robotics.

Human-only motion generation \cite{zhang2024motiondiffuse, tevet2023human, zhang2023generating, karunratanakul2023guided, chen2023executing, guo2024momask, zhou2025emdm} primarily focuses on learning human behavior patterns in isolation, without taking the environmental context into account. However, human motion is inherently carried out through interactions with objects, other individuals, and the surrounding environment. In real-world scenarios, human actions are are typically responses to external stimuli, which highlights the essential role of interaction in shaping human behavior. Unlike traditional motion generation, which overlooks these complex dynamics, interaction generation involves the simultaneous changes in both human motion and the movement of corresponding interactive entities. This dual consideration facilitates more meaningful and contextually appropriate behaviors, reflecting the dynamic nature of real-world interactions.

\begin{figure}[h!]
	\centering
	\includegraphics[width=\linewidth]{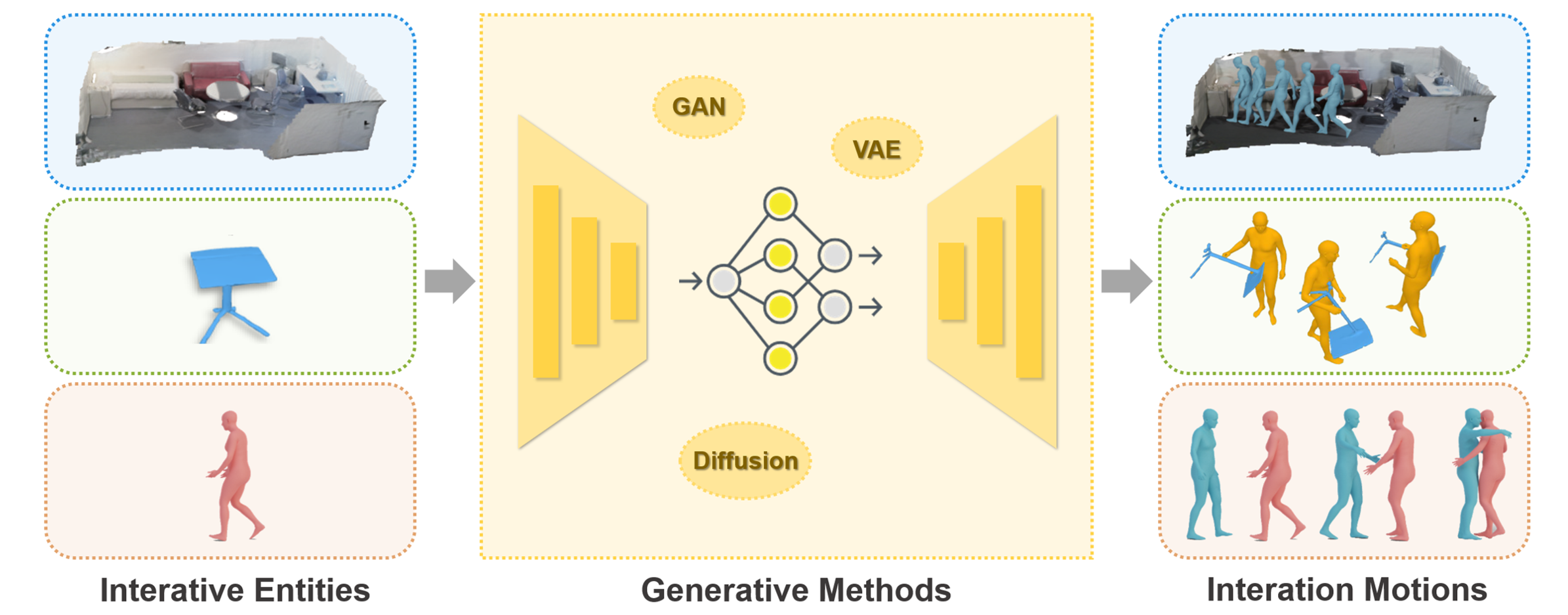}
	\caption{\label{fig:approach}
		An overview of typical human interaction generation approaches. Example figures from \cite{huang2023diffusion, li2023object, xu2024inter}.}
\end{figure}

\begin{figure}[h!]
	\centering
	\includegraphics[width=0.95\linewidth]{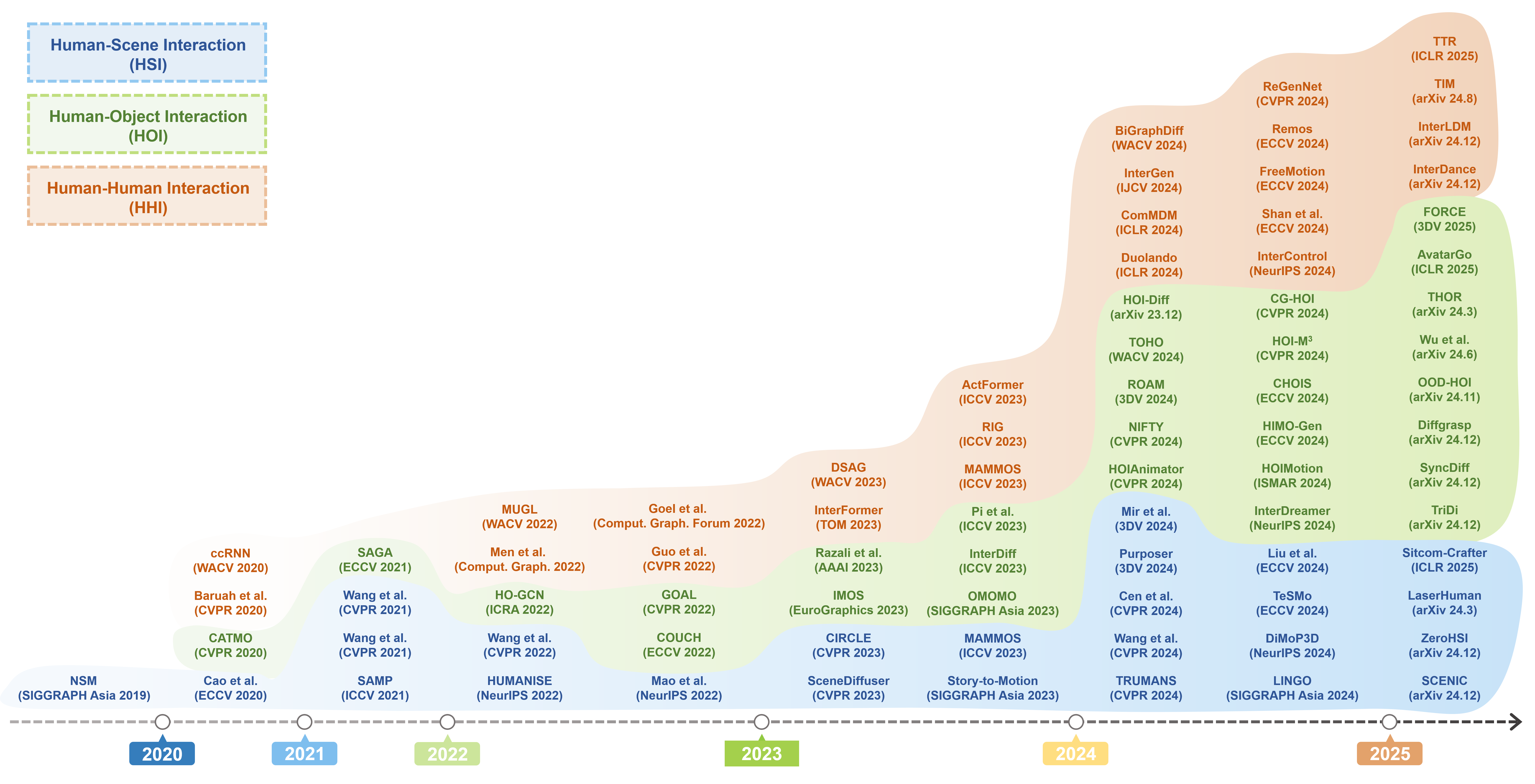}
	\caption{\label{fig:timeline}
		Recent advances of human interaction generation methods with different interactive entities.}
\end{figure}

To accomplish human interaction generation, there are a few fundamental technologies that greatly promote the accuracy and realism of generated movements. Various 3D modeling methods offers different representation forms for both interactive human \cite{loper2023smpl, MANO, SMPL-X} and interactive entities \cite{wang20243d}, which enhance the motion data quality and contact relationship accuracy of human interaction. And the motion capture methods \cite{vantagecutting, zhang2012microsoft, paulich2018xsens} provide the possibility to record the interaction motion and correlation of all participants simultaneously. The remarkable process of generative models lead to the prospective success in artificial intelligence generated contents including texts \cite{brown2020language, ouyang2022training}, images \cite{karras2020analyzing, karras2021alias}, videos \cite{ho2022video, ho2022imagen}, and 3D objects \cite{luo2021diffusion, qian2023magic123}. Typically, Generative Adversarial Networks (GAN) \cite{goodfellow2014generative}, Variational Autoencoders (VAEs) \cite{kingma2013auto} and Diffusion Models \cite{ho2020denoising} are the generative methods that most commonly used in human interaction generation.

Though going through rapid developing progress, human interaction generation faces several key challenges. First, the absence of large approachable datasets limit the interactive entities into few categories \cite{zhang2024force, petrov2024tridi, zhang2024hoi}. And available datasets lack natural language annotations that express interactive intention, leaving the motion uncontrollable \cite{li2024zerohsi, li2025controllable}. The short of high quality datasets results in bad robustness and generalization of human interactions, posing great challenges in dynamic and unpredictable environments. Second, the physical plausibility of human motion and the contact accuracy between human and interactive entities often lead to artificial effects \cite{he2024syncdiff, zhang2024diffgrasp, cong2024laserhuman, jiang2024autonomous}. Data-driven generation methods simulate the distributions of training datasets, generating seemingly plausible results but failing to understand physical properties such as resistance, mass, and velocity \cite{zhang2024force, cong2024laserhuman, braun2024physically, shimada2024macs}. As a result, these methods often produce distorted phenomena like sliding, penetration, and floating, as they cannot generate results that fully comply with physical laws. Additionally, the controllability of human interaction generation needs further explored. Unlike isolated human motion generation, interaction generation are primary constrained by the static contribute and dynamic characteristic of interactive entities. The additional constraint of controllable signals such as textual description \cite{guzov2021human}, action style \cite{jang2022motion, xu2020hierarchical}, human emotion \cite{tan2023emmn}, and fine-grained process control \cite{zhang2023finemogen} bring new burdens to the generation. Addressing these challenges is crucial for enhancing the accuracy, realism, and applicability of human interaction.

With the rapid development of generative tasks, human interaction generation has attracted widespread attention, leading to significant advancements in the field. This paper summarizes data-driven human interaction methods in recent years. Figure \ref{fig:approach} and Figure \ref{fig:timeline} illustrate typical approaches and the developmental timeline of human interaction generation. Section \ref{Scope} defines the scope of works considered in this survey. In section \ref{Preliminaries}, we provide preliminary technologies of human interaction generation, including the representation and capture methods of motion data. Section \ref{Human-Scene Interaction}, \ref{Human-Object Interaction}, and \ref{Human-Human Interaction} surveys three different categories of human interaction generation based on the interactive entities: human-object interaction, human-scene interaction, and human-human interaction. In section \ref{Dataset and Evaluation Metrics}, commonly used datasets and evaluation metrics are introduced. Last, a conclusion and future potential directions for human interaction generation are discussed in Section \ref{Conclusion and Future Work}.

\section{Scope} \label{Scope}
This survey primarily focuses on deep learning approaches for generating human interaction. We categorize interactive entities into three types: scene, object and other human. Methods for generating isolated human motions \cite{zhang2024motiondiffuse, tevet2023human, zhang2023generating, karunratanakul2023guided, chen2023executing, guo2024momask, zhou2025emdm} are excluded, as this survey is specifically concerned with approaches that generate interaction motions within a contextual environment. While reinforcement learning methods \cite{chao2021learning, bae2023pmp, hassan2023synthesizing, wang2023physhoi, zhao2023synthesizing, xiao2024unified, pan2024synthesizing} that leverage physical simulation environments can also generate task-oriented human interactions, only data-driven generation methods are considered in the scope of this survey. Furthermore, interaction pose generation \cite{zhang2020generating, hassan2021populating, li2019putting, zhao2022compositional, zheng2023coop, tendulkar2023flex}, trajectory generation \cite{rossi2021human, zhang2022wanderings, jiang2023continuous}, and hand-object interaction generation \cite{zhang2021manipnet, taheri2023grip, zhou2024gears, li2024clickdiff, liu2024geneoh} are excluded, as this survey is dedicated to works that generate sequential, whole-body human interaction motions.

\section{Preliminaries} \label{Preliminaries}
Interaction motion involves two essential parts: the interacting human and the interactive entities. The interacting human represents the individual who participates in the interaction and leads the movements, while the interactive entities consist of any objects or individuals within the environment that respond to the human’s actions. We first separately introduce the 3D model representation and  motion capture process of this two participants, then discuss the generation methods commonly used in human interaction generation.
\subsection{3D Model Representation}
\subsubsection{Human Representation}
\textbf{Skeletal-based representation.} Skeletal-based human representation models 3D human figure from the perspective of anatomy, simplify human body to skeleton and joints. Following the natural hierarchy of human body, bones are connected to others via joints, allowing for realistic articulation. Human motion is represented by a sequence of human poses and human pose is represented by kinetic attributes of joints or bones, such as position, velocity, and rotation. A typical skeletal-based human representation \cite{mandery2015kit} is organized by a hierarchical structure, with the root joint, usually the pelvis, positioned in the global coordinate system. The root joint serves as the anchor of a skeleton, defining the overall position and orientation of the human body in 3D space. From the root joint, all other joints are sequentially linked throughout the skeletal hierarchy, defined in local coordinates relative to their parent joints. To transfer motion capture data to a skeletal-based representation, animation experts or deep learning methods rig the recorded human surface to the skeleton model, enabling the mapping between the captured human body surface and the digital human skeleton.

Skeletal-based human representation provides real-time locomotion and the ability to animate complex movements with high fidelity. However, skeletal-based representation lacks the capacity of describing body shape and geometry, failing to provide detailed information about changes in the human body during interactions.

\textbf{Parametric-based representation.} Parametric-based human representation describes human body from statistical aspect, using parameterized function to model human body mesh. Parametric-based representation models decouple human mesh to limited couples of parameters which describes different aspects of body geometry feature. Driven by large-scale scanned data, models capture human representation distribution in low-dimensional space and compresses the complexity. Skinned Multi-Person Linear (SMPL) model \cite{loper2023smpl} is the most widely used parametric-based human representation. The SMPL model decomposes the 3D human body into shape-dependent and pose-dependent deformations. The shape parameters define features such as height, weight, and body proportions, while the pose parameters control joint rotations and the overall body posture. Subsequent works, such as STAR \cite{shi2021star}, SMPL-H \cite{MANO}, and SMPL-X \cite{SMPL-X}, have continuously extended the SMPL model to create more refined representations that include not only the body, but also the head and hands. Parametric-based human representation relies on high-quality data, including scanned meshes, depth maps, RGB images, and other modalities. After data preprocessing, key human features are extracted and organized into parameter categories. These features are further optimized through statistical models and machine learning techniques to learn representative parameters that capture human shape and pose variations.

Leveraging low-dimensional parameters to represent the high-dimensional human body enhances the efficiency of the representation. Modeling the human mesh allows for better shape and geometric fidelity compared to skeletal-based human representations. However, parametric-based representations heavily depend on large-scale, high-quality data, which limits their development and makes them less flexible in capturing extreme shapes.

\subsubsection{Interactive Entities Representation}
\begin{figure}[h!]
	\centering
	\includegraphics[width=\linewidth]{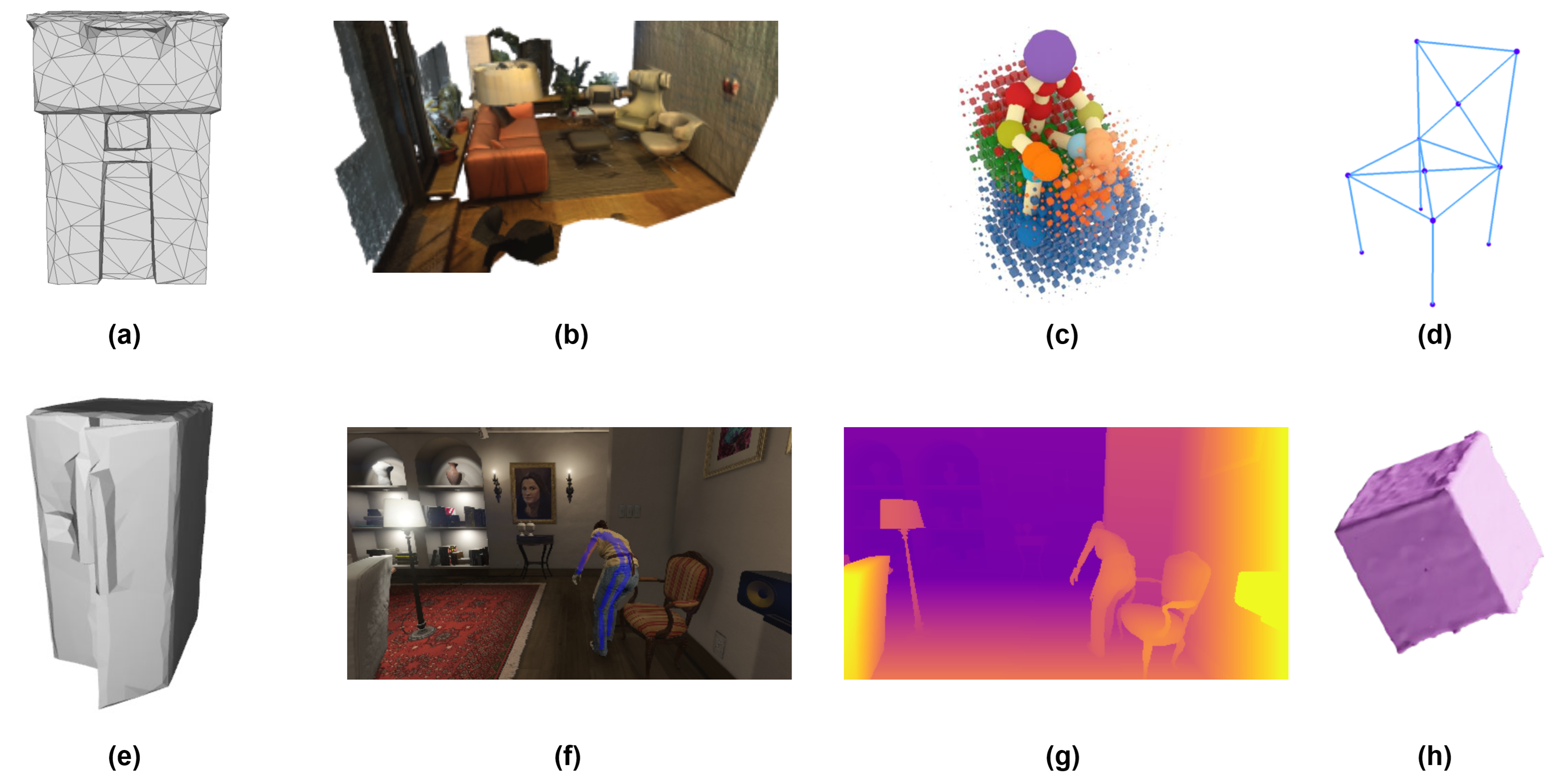}
	\caption{\label{fig:representation}
		Interactive entities representations of (a) mesh \cite{xu2021d3d}, (b) point cloud \cite{wang2022humanise}, (c) voxel \cite{savva2016pigraphs}, (d) keypoints \cite{wan2022learn}, (e) computer-aided design \cite{xu2021d3d}, (f) RGB images \cite{cao2020long}, (g) RGB-D images \cite{cao2020long}, and (h) signed distance function \cite{li2023object}.}
\end{figure}

Interactive entities can be represented in various ways, describing them in terms of both geometric properties and visual attributes, as shown in Figure \ref{fig:representation}.

\textbf{Mesh.} Mesh describes the surface of an object through vertices, edges, and faces, making it the most widely used representation for interactive objects \cite{wang2021synthesizing, wang2022towards, taheri2020grab, bhatnagar2022behave, zhang2024core4d}. Meshes can be directly obtained using 3D scanners, providing detailed surface geometry. However, processing highly detailed meshes can be computationally intensive, especially for dynamic interactions.

\textbf{Point Cloud.} Point cloud represent a set of discrete points in 3D space, offering a straightforward depiction of scene geometry \cite{cong2024laserhuman, wang2022humanise, qi2017pointnet, araujo2023circle}. Point clouds are extensively used in human-scene interaction task as they can be directly captured using sensors such as LiDAR or depth cameras. Despite their simplicity, point clouds are inherently static and lack connectivity information, which makes simulating dynamic interactions challenging.

\textbf{Voxel.} Voxel divides 3D space into a regular grid of cubic units, with each voxel containing information about the space it occupies \cite{jiang2024autonomous, wu20153d, hassan2021stochastic}. Unlike meshes and point clouds, voxels provide a volumetric representation that captures both the internal and external structures of an object or environment. This makes them particularly suitable for collision detection, but their resolution is limited by computational and memory constraints.

\textbf{Keypoints.} Keypoints represent an object through a sparse set of discrete points corresponding to significant structures, such as corners or centers of surfaces \cite{jiang2024autonomous, wan2022learn}. They are computationally efficient for representing regular shapes and are commonly used in simplified interaction modeling. However, their abstract nature often lacks the detail required to model complex interactions effectively.

\textbf{Computer-Aided Design (CAD).} CAD models provide parametric, highly precise representations explicitly defined by their geometric structures and attributes \cite{xu2021d3d, liu2023interactive}. They are editable and often derived from existing design datasets, making them ideal for structured applications. However, CAD models are typically static and predefined, limiting their applicability in dynamic interaction contexts.

\textbf{RGB Images.} RGB images are 2D projections of scenes where each pixel encodes explicit color information in terms of the red, green, and blue color channels \cite{cao2020long, wang2021scene, kim2024parahome}. They lack depth information and 3D spatial context, making it difficult to accurately understand the geometry and spatial relationships in a scene.

\textbf{RGB-D Images.} RGB-D images extend RGB images by incorporating depth information for each pixel, providing an explicit description of both the visual and spatial properties of the environment \cite{cao2020long, Matterport3D}.

\textbf{Signed distance function (SDF).} SDF represents the geometry of an object by defining the distance of any point in space to the nearest surface \cite{li2023object, chen2024sitcom}. Positive values denote points outside the object, negative values represent points inside, and zero corresponds to points lying exactly on the surface. As an implicit representation, the SDF offers strong generalization capabilities, allowing it to encode the fundamental geometric patterns of objects and interactions. This makes it particularly useful for models that aim to infer and generalize interaction patterns across a wide variety of objects and environments.

\subsection{Motion Capture}
\begin{figure}[h!]
	\centering
	\includegraphics[width=\linewidth]{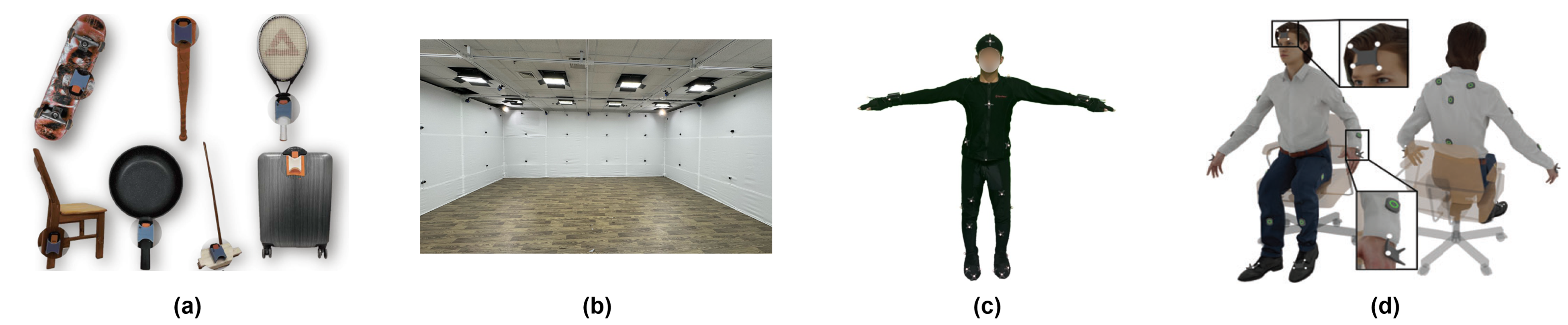}
	\caption{\label{fig:mocap}
		Motion capture techniques. (a) Examples of IMUs-based mocap system with inertial sensors attached to the entities' surface \cite{zhao2024m}. (b) Examples of markerless mocap system \cite{zhang2024hoi}, which uses cameras and algorithms to track human motion. (c) Examples of marker-based mocap system \cite{xu2024inter}, where reflective markers are attached to the subject. (d) Examples of hybrid mocap system \cite{jiang2022chairs}, combining both optical and inertial setups.}
\end{figure}

Motion capture techniques simultaneously record human interaction motions and the corresponding changes in interactive entities, providing a fundamental basis for data-driven human interaction generation. Examples of motion capture are shown in Figure \ref{fig:mocap}.

\textbf{Inertial-based motion capture.} Inertial-based motion capture systems \cite{xsens, noitom} employ wearable inertial measurement units (IMUs) attached to the human body or objects, without relying on external cameras or markers. This method offers high portability and flexibility, enabling operation across diverse environments. However, it may drift over time and requires periodic calibration to ensure accuracy. Additionally, motion capture gloves embedded with IMUs are included in this category, allowing for the precise capture of detailed hand and finger movements in human motion.

\textbf{Optical-based motion capture.} Optical-based motion capture systems employ one or more cameras to capture visual data and reconstruct 3D motions. These systems are generally categorized into marker-based \cite{vicon, optitrack, nokov} and markerless \cite{kinect} approaches. Marker-based systems rely on reflective markers attached to the subject’s body, which are tracked by synchronized cameras to achieve high precision. Advanced approaches such as Mosh and Mosh++ \cite{loper2014mosh} integrate parametric human models with sparse marker data to reconstruct both detailed body shapes and motions, further enhancing the expressiveness of marker-based methods. However, they require a controlled environment for optimal performance, limiting their applicability in outdoor or uncontrolled settings. In contrast, markerless approaches leverage computer vision algorithms on RGB or RGB-D data to estimate poses without the need for external markers, providing greater convenience and flexibility. Despite these advantages, markerless methods are generally less robust to variations in lighting conditions and occlusions.

\textbf{Hybrid motion capture.} Hybrid motion capture systems combine multiple sensing modalities to leverage the complementary advantages of optical-based and inertial-based approaches \cite{xu2024inter, cong2024laserhuman, lv2024himo, li2024interdance}. Inertial sensors provide reliable local orientation data and are unaffected to occlusions, while optical tracking offers precise global position references with high spatial accuracy. Hybrid motion capture systems effectively mitigate issues such as substantial drift and environmental constraints, enabling their use in diverse scenarios, including expansive outdoor scenarios and cluttered indoor environments. However, achieving accurate alignment and synchronization of motion data across different modalities remains a significant challenge.

\subsection{Generation Methods}
\begin{figure}[h!]
	\centering
	\includegraphics[width=\linewidth]{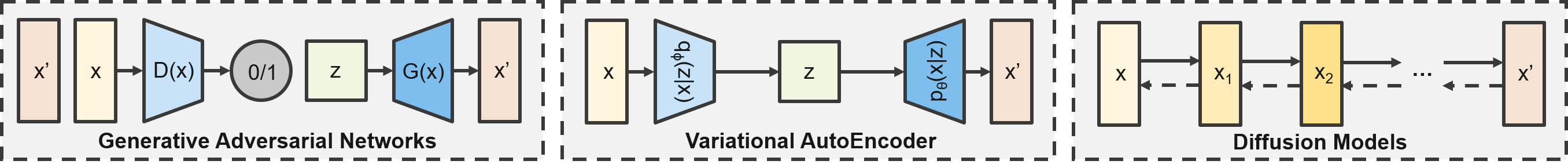}
	\caption{\label{fig:generative}
		Commonly used generative models in human interaction generation.}
\end{figure}

Following the exploration of prior works in isolated human motion generation, human interaction motion generation leverages similar foundational generative models such as Generative Adversarial Networks, Variational Autoencoders and Diffusion Models to handle interactive scenarios, as shown in Figure \ref{fig:generative}. These models integrate multi-modal inputs, such as spatial configurations, object properties, and textual descriptions to enhance the realism and contextually appropriateness. In the following, we will introduce how each of these generative models is applied in the context of human interaction generation.

\textbf{Generative Adversarial Networks.} Generative Adversarial Networks (GANs) \cite{goodfellow2014generative}  are a class of generative models that consist of two neural networks, the generator $G$ and the discriminator $D$, which are trained simultaneously in a competitive framework. The generator $G$ aims to produce data $x_G$ that resembles ground truth data $x$, while the discriminator $D$ seeks to distinguish between ground truth data $x \sim p_{data}(x)$ and generated data$x_G=G(z)$, where $z \sim p_{z}(z)$ is a random noise vector sampled from a prior distribution. This adversarial process is formulated as a minimax optimization problem, where the generator tries to minimize the discriminator's ability to distinguish generated samples from real ones, while the discriminator attempts to maximize its classification performance. The loss function of the discriminator $D$ and the generator $G$ are defined as:
\begin{equation}\label{eq:LD}
	\mathcal{L}_D = - \mathbb{E}_{x \sim p_{data}(x)} [\log D(x)] - \mathbb{E}_{z \sim p_z} [\log (1 - D(G(z)))]
\end{equation}
\begin{equation}\label{eq:LG}
	\mathcal{L}_G = - \mathbb{E}_{z \sim p_z} [\log D(G(z))]
\end{equation}
GANs facilitates the generation of varied content by learning intricate patterns from vast datasets. However, GANs also come with several drawbacks. One of the most significant challenges is their training instability, often requiring careful tuning of hyper-parameters and extensive computational resources. Issues like mode collapse and vanishing gradients also hinder the diversity and richness of the generated content. While GANs perform well on spatial patterns, maintaining long-term temporal consistency in motions remains a challenge. Moreover, GANs require large amounts of high-quality, annotated data for training. Continued advancements in GAN architectures such as DCGAN \cite{radford2015unsupervised}, WGAN \cite{arjovsky2017wasserstein}, CycleGAN \cite{zhu2017unpaired}, and StyleGAN \cite{karras2019style} are designed to overcome these limitations and make the best of these generative models.

\textbf{Variational Autoencoders.} Variational Autoencoders (VAEs) \cite{kingma2013auto} consists of an encoder $q_{\phi}(\mathbf{z}|\mathbf{x})$, which maps the input data $x$ to a latent distribution $z$, and a decoder $p_{\theta}(\mathbf{x}|\mathbf{z})$, which reconstructs the input from the latent space. To ensure the latent representation follows a continuous and regularized distribution, VAEs introduce a Kullback-Leibler (KL) divergence penalty that aligns the posterior $q_{\phi}(\mathbf{z}|\mathbf{x})$ with a predefined prior $p(z)$. The loss function is defined as:
\begin{equation}\label{eq:LG}
\mathcal{L}_{VAE} = \mathbb{E}_{q_{\phi}(\mathbf{z}|\mathbf{x})} \left[ \log p_{\theta}(\mathbf{x}|\mathbf{z}) \right] - D_{KL} \left( q_{\phi}(\mathbf{z}|\mathbf{x}) || p(\mathbf{z}) \right)
\end{equation}
VAEs excel in learning structured and continuous latent representations of data, which facilitates smooth interpolation and meaningful manipulation of generated content. Additionally, VAEs are known for their training stability, ensuring consistent performance during the training process. Their scalability allows them to efficiently handle large and complex datasets. However, VAEs also face notable challenges. Due to the inherent trade-off between the reconstruction accuracy and the regularization imposed by the variational framework, the generated samples of VAEs often lack the fine-grained detail and sharpness achievable by other generative models like GANs. Moreover, VAEs can suffer from issues such as posterior collapse, where the latent variables fail to capture meaningful information, leading to less expressive representations and reduced diversity. Various improvements and extensions such as Conditional Variational Autoencoders (CVAEs) \cite{sohn2015learning} and Vector Quantized VAEs (VQ-VAEs) \cite{van2017neural} have been developed to enhance expressiveness and the fidelity of VAEs.

\textbf{Diffusion Models.} Diffusion models \cite{ho2020denoising} have recently emerged as a powerful generative framework for high-quality content generation, leveraging a forward diffusion process and a reverse denoising process. The forward diffusion process gradually adds Gaussian noise to the data $x \sim p_{data}(x)$, progressively transforming it into pure noise. This process is formally defined as:
\begin{equation}\label{eq:LG}
q(x_t | x_{t-1}) = \mathcal{N}(x_t; \sqrt{1 - \beta_t} x_{t-1}, \beta_t I)
\end{equation}
where ${x}_t$ is the noised sample at timestep $t$, $\beta_t$ represents the variance schedule controlling the noise level, and $I$ is the identity matrix. The reverse denoising process aims to iteratively remove the added noise, recovering realistic data by approximating the reverse conditional distribution using a parameterized neural network $\epsilon_\theta$. This reverse process is defined as:
\begin{equation}\label{eq:LG}
p_{\theta}(x_{t-1} | x_t) = \mathcal{N}(x_{t-1}; \mu_\theta(x_t, t), \Sigma_\theta(x_t, t))
\end{equation}
where $\mu_\theta(x_t, t)$ and $\Sigma_\theta(x_t, t)$ denote the predicted mean and variance of the denoised distribution. During training, the network is optimized to minimize the difference between the predicted noise $\epsilon_\theta(x_t, t)$ and the actual noise.

Diffusion models have emerged as a leading generative framework generation tasks due to their ability to produce high-quality, diverse, and controllable outputs.
However, the iterative sampling process of diffusion models require a large number of steps, leading to slow inference speed and high computational cost, which hinders their development in real-time applications.
Diffusion models have emerged as a leading generative framework for content generation tasks due to their ability to produce high-quality, diverse, and controllable outputs. However, their iterative sampling process requires a large number of steps, resulting in slow inference speed and high computational costs, which limit their deployment in real-time applications. Several improved versions have been proposed to enhance the efficiency of diffusion models, such as Denoising Diffusion Implicit Models (DDIM) \cite{song2020denoising} and Latent Diffusion Models (LDM) \cite{rombach2022high}.

\section{Human-Scene Interaction} \label{Human-Scene Interaction}
The human-scene interaction generation task aims to generate realistic and smooth human motions that interact naturally with elements within a given scene, emphasizing goal achievement, obstacle avoidance, and semantic accuracy in the context of the environment.

\subsection{Human-Scene Interaction Generation}
Several researchers explored the possibilities of human interaction with static scenes based on simple scene conditions and human initial states.
NSM \cite{starke2019neural} first introduces a CNN-LSTM-based framework for generating human-scene interactions using bi-directional control and volumetric geometry, enabling real-time adaptability to diverse 3D environments.
Wang et al. \cite{wang2021scene} propose a scene-aware generative network based on GAN, which decouples human motion synthesis into trajectory generation and pose generation while employing geometry-aware discriminators to ensure compatibility with scene semantics and structures.
As the field progressed, approaches advanced towards more flexible and context-aware models that integrate a broader range of information, such as goal position and orientation.
Similarly, Wang et al. \cite{wang2021synthesizing} introduce a hierarchical two-stage framework that uses CVAE to generate static human body poses at subgoal locations and bi-directional LSTM to synthesize short-term motion sequences between these subgoals, incorporating geometry-aware optimization to ensure physical plausibility and natural interactions in complex scenes.
CIRCLE \cite{araujo2023circle} leverages a Transformer-based generative model that first initializes human motion sequences through interpolation and then refines them using scene-conditioned attention mechanisms, leveraging PointNet++ \cite{qi2017pointnet++} to encode scene geometry and synthesize physically plausible scene-aware human motions.
SceneDiffuser \cite{huang2023diffusion} leverages the iterative denoising process of diffusion models to integrate scene-aware generation, physics-based optimization, and goal-oriented planning into a unified framework, addressing the challenges of posterior collapse and module inconsistencies in prior methods.
Mir et al. \cite{mir2024generating} propose a framework utilizing a Transformer-based WalkNet for long range path following and a masked autoencoder-based TransNet for smooth transitions between actions, enabling continual human motion synthesis in diverse 3D scenes without scene-specific training data.
Liu et al. \cite{liu2024revisit} introduce an autoregressive controller trained on their motion occupancy database, which enables collision-free generation of static human-scene interactions by modeling human-occupancy dynamics without relying on explicit 3D scene data.

\renewcommand\arraystretch{1}
\setlength\tabcolsep{2pt}
\setlength{\LTcapwidth}{\textwidth}
\setlength{\LTleft}{0pt}
\setlength{\LTright}{0pt}

\scriptsize
\onecolumn
\begin{longtable}{
    >{\centering\arraybackslash}m{0.03\textwidth}
    >{\centering\arraybackslash}m{0.16\textwidth}
    >{\centering\arraybackslash}m{0.18\textwidth}
    >{\centering\arraybackslash}m{0.14\textwidth}
    >{\centering\arraybackslash}m{0.17\textwidth}
    >{\centering\arraybackslash}m{0.04\textwidth}
    >{\centering\arraybackslash}m{0.05\textwidth}
    >{\centering\arraybackslash}m{0.16\textwidth}
}
    \caption{Representative works of human interaction generation. In the \textbf{Text} column, "T" represents Text, "A" represents Action Label, "M" represents Music, "N" represents None. In the \textbf{Parts} column, "B" represents Body, "H" represents Hand, "F" represents Face. In the \textbf{Highlight} column, the meaning varies by task: it represents scene representation for Human Scene Interaction, object motion for Human-Object Interaction, and the number of humans for Human-Human Interaction.} \\
    \toprule[0.4mm]
    \addlinespace[0.2mm]
    \cmidrule[0.6pt]{1-8}
    & \textbf{Method} & \textbf{Venue} & \textbf{Model} & \textbf{Dataset} & \textbf{Text} & \textbf{Parts} & \textbf{Highlight} \\
    \endfirsthead

    \endfoot
    \endlastfoot
    
      \midrule
      \addlinespace[1.0pt]
      \multirow{25}{*}{\makecell{\rotatebox{90}{\textbf{Human-Scene Interaction (HSI)}}}}
      & NSM \cite{starke2019neural} & SIGGRAPH Asia 2019 & CNN, LSTM & \cite{starke2019neural} & N & B & Votex \\
      \noalign{\global\renewcommand{\arraystretch}{1}}
      
      \rowcolor{myLightBlue}\omit & Cao et al. \cite{cao2020long} & ECCV 2020 & VAE, transformer & \cite{cao2020long, hassan2019resolving} & N & B & RGB Image \\
      & Wang et al. \cite{wang2021scene} & CVPR 2021 & GAN & \cite{cao2020long, hassan2019resolving} & N & B & RGB Image \\
      \rowcolor{myLightBlue}\omit & Wang et al. \cite{wang2021synthesizing} & CVPR 2021 & VAE, LSTM &  \cite{Matterport3D, hassan2019resolving} & N & BH & Mesh \\
      & SAMP \cite{hassan2021stochastic} & ICCV 2021 & VAE & \cite{hassan2021stochastic} & A & B & Votex \\
      \rowcolor{myLightBlue}\omit & Wang et al. \cite{wang2022towards} & CVPR 2022 & VAE, Transformer &  \cite{Matterport3D, hassan2019resolving} & A & B & Mesh \\
      & HUMANISE \cite{wang2022humanise} & NeurIPS 2022 & VAE & \cite{wang2022humanise} & T & B & Point Cloud \\
      \rowcolor{myLightBlue}\omit & Mao et al. \cite{mao2022contact} & NeurIPS 2022 & CNN, GRU & \cite{cao2020long, hassan2019resolving} & N & B & Point Cloud \\
      & CIRCLE \cite{araujo2023circle} & CVPR 2023 & Transformer & \cite{araujo2023circle} & N & BHF & Point Cloud \\
      \rowcolor{myLightBlue}\omit & SceneDiffuser \cite{huang2023diffusion} & CVPR 2023 & Diffusion &  \cite{hassan2019resolving} & N & B & Point Cloud \\
      & MAMMOS \cite{lim2023mammos} & ICCV 2023 & VAE & \cite{hassan2019resolving, replica19arXiv} & A & BH & CAD/Mesh/ Point Cloud \\
      \rowcolor{myLightBlue}\omit & Story-to-Motion \cite{qing2023story} & SIGGRAPH Asia 2023 & LLM, Transformer & \cite{AMASS:2019} & T & B & - \\
      & Mir et al. \cite{mir2024generating} & 3DV 2024 & Transformer & \cite{guzov2021human, Matterport3D, replica19arXiv, AMASS:2019, dai2017scannet} & N & B & Point Cloud \\
      \rowcolor{myLightBlue}\omit & Purposer \cite{ugrinovic2024purposer} & 3DV 2024 & Discrete AR & \cite{wang2022humanise, hassan2019resolving, BABEL:CVPR:2021} & A & B & Point Cloud \\
      & Cen et al. \cite{cen2024generating} & CVPR 2024 & LLM, Diffusion & \cite{wang2022humanise, hassan2019resolving} & T & B & Point Cloud \\
      \rowcolor{myLightBlue}\omit & Wang et al. \cite{wang2024move} & CVPR 2024 & Diffusion & \cite{wang2022humanise, Guo_2022_CVPR} & T & B & Point Cloud \\
      & TRUMANS \cite{jiang2024scaling} & CVPR 2024 & Diffusion & \cite{zhang2020generating, taheri2020grab, jiang2024scaling} & A & BH & Point Cloud \\         
      \rowcolor{myLightBlue}\omit & Liu et al. \cite{liu2024revisit} & ECCV 2024 & Transformer & \cite{liu2024revisit} & N & B & Votex \\
      & Tesmo \cite{yi2024generating} & ECCV 2024 & Diffusion & \cite{hassan2021stochastic, Guo_2022_CVPR, fu20213d} & T & B & BPS/Floor Maps \\
      \rowcolor{myLightBlue}\omit & DiMoP3D \cite{lou2024harmonizing} & NeurIPS 2024 & Diffusion & \cite{araujo2023circle, zheng2022gimo} & N & B & Point Cloud \\    
      & LINGO \cite{jiang2024autonomous} & SIGGRAPH Asia 2024 & Diffusion & \cite{jiang2024autonomous} & T & B & Votex \\    
      \rowcolor{myLightBlue}\omit & LaserHuman \cite{cong2024laserhuman} & arXiv 24.3 & Diffusion & \cite{cong2024laserhuman} & T & B & Point Cloud \\    
      & ZeroHSI \cite{li2024zerohsi} & arXiv 24.12 & Framework & - & T & B & 3DGS \\     
      \rowcolor{myLightBlue}\omit & SCENIC \cite{zhang2024scenic} & arXiv 24.12 & Diffusion & \cite{hassan2021stochastic, zhang2024scenic} & T & B & BPS/Distance Field \\    
      & Sitcom-Crafter \cite{chen2024sitcom} & ICLR 2025 & Framework & \cite{xu2024inter, replica19arXiv, liang2024intergen} & T & BH & SDF \\

      \addlinespace[-1.0pt]
      \midrule
      \addlinespace[1.0pt]
      \multirow{29}{*}{\makecell{\rotatebox{90}{\textbf{Human-Object Interaction (HOI)}}}}
      & CATMO \cite{corona2020context} & CVPR 2020 & GNN, RNN & \cite{10.1177/0278364913478446} & N & B & yes \\
      \noalign{\global\renewcommand{\arraystretch}{1}}

      \rowcolor{myLightBlue}\omit & SAGA \cite{wu2022saga} & ECCV 2021 & VAE & \cite{taheri2020grab} & N & BH & no \\
      & HO-GCN \cite{wan2022learn} & ICRA 2022 & GCN & \cite{wan2022learn} & A & B & yes \\
      \rowcolor{myLightBlue}\omit & GOAL \cite{taheri2022goal} & CVPR 2022 & VAE & \cite{taheri2020grab} & N & BHF & no \\
      & COUCH \cite{zhang2022couch} & ECCV 2022 & LSTM, VAE & \cite{zhang2022couch} & N & B & no \\      
      \rowcolor{myLightBlue}\omit & Razali et al. \cite{razali2023action} & AAAI 2023 & GNN, GRU & \cite{mandery2015kit} & A & BH & yes \\
      & IMOS \cite{ghosh2023imos} & EuroGraphics 2023 & VAE & \cite{taheri2020grab} & A & BHF & yes \\
      \rowcolor{myLightBlue}\omit & Pi et al. \cite{pi2023hierarchical} & ICCV 2023 & VAE, Diffusion & \cite{hassan2021stochastic, starke2019neural, zhang2022couch} & A & B & no \\
      & InterDiff \cite{xu2023interdiff} & ICCV 2023 & Diffusion & \cite{bhatnagar2022behave} & N & B & yes \\
      \rowcolor{myLightBlue}\omit & OMOMO \cite{li2023object} & SIGGRAPH Asia 2023 & Diffusion & \cite{li2023object} & N & B & yes \\
      & HOI-Diff \cite{peng2023hoi} & arXiv 23.12 & Diffusion & \cite{bhatnagar2022behave} & T & B & yes \\
      \rowcolor{myLightBlue}\omit & TOHO \cite{li2024task} & WACV 2024 & VAE & \cite{taheri2020grab} & A & BH & yes \\
      & ROAM \cite{zhang2024roam} & 3DV 2024 & NSM & \cite{zhang2024roam} & N & B & no \\
      \rowcolor{myLightBlue}\omit & NIFTY \cite{kulkarni2024nifty} & CVPR 2024 & Diffusion & \cite{bhatnagar2022behave, rempe2021humor} & A & B & no \\
      & HOIAnimator \cite{song2024hoianimator} & CVPR 2024 & Diffusion & \cite{bhatnagar2022behave, huang2022intercap} & T & B & yes \\    
      \rowcolor{myLightBlue}\omit & CG-HOI \cite{diller2024cg} & CVPR 2024 & Diffusion & \cite{bhatnagar2022behave, jiang2022chairs} & T & B & yes \\
      & HOI-M\textsuperscript{3} \cite{zhang2024hoi} & CVPR 2024 & Diffusion & \cite{zhang2024hoi} & N & B & yes \\      
      \rowcolor{myLightBlue}\omit & CHOIS \cite{li2025controllable} & ECCV 2024 & Diffusion & \cite{li2023object} & T & B & yes \\
      & HIMO-Gen \cite{lv2024himo} & ECCV 2024 & Diffusion & \cite{lv2024himo} & T & BH & yes \\
      \rowcolor{myLightBlue}\omit & HOIMotion \cite{hu2024hoimotion} & ISMAR 2024 & GCN & \cite{pan2023aria, kratzer2020mogaze} & N & B & no \\
      & InterDreamer \cite{xu2024interdreamer} & NeurIPS 2024 & Framework & - & T & B & yes \\
      \rowcolor{myLightBlue}\omit & THOR \cite{wu2024thor} & arXiv 24.3 & Diffusion & \cite{bhatnagar2022behave} & T & B & yes \\
      & Wu et al. \cite{wu2024human} & arXiv 24.6 & Framework & \cite{li2023object, taheri2020grab, Guo_2022_CVPR} & T & BH & yes \\    
      \rowcolor{myLightBlue}\omit & OOD-HOI \cite{zhang2024ood} & arXiv 24.11 & Diffusion & \cite{taheri2020grab} & T & BH & yes \\
      & Diffgrasp \cite{zhang2024diffgrasp} & arXiv 24.12 & Diffusion & \cite{taheri2020grab, fan2023arctic} & N & BH & yes \\
      \rowcolor{myLightBlue}\omit & SyncDiff \cite{he2024syncdiff} & arXiv 24.12 & Diffusion & \cite{taheri2020grab, zhang2024core4d, liu2024taco, zhan2024oakink2} & A & BH & yes \\
      & TriDi \cite{petrov2024tridi} & arXiv 24.12 & Diffusion & \cite{li2023object, taheri2020grab, bhatnagar2022behave, huang2022intercap} & T & BHF & yes \\
      \rowcolor{myLightBlue}\omit & FORCE \cite{zhang2024force} & 3DV 2025 & VAE & \cite{zhang2024force} & A & B & yes \\
      & AvatarGO \cite{cao2024avatargo} & ICLR 2025 & Framework & - & T & BHF & yes \\

      \addlinespace[-1.0pt]
      \midrule
      \addlinespace[1.0pt]
      \multirow{24}{*}{\makecell{\rotatebox{90}{\textbf{Human-Human Interaction (HHI)}}}}
      & ccRNN \cite{kundu2020cross} & WACV 2020 & RNN & \cite{kundu2020cross, yun2012two} & N & B & 2 \\
      \noalign{\global\renewcommand{\arraystretch}{1}}

      \rowcolor{myLightBlue}\omit & Baruah et al. \cite{baruah2020multimodal} & CVPR 2020 & RNN & \cite{yun2012two, hu2013efficient} & A & B & 2 \\
      & MUGL \cite{maheshwari2022mugl} & WACV 2022 & VAE & \cite{liu2019ntu} & A & B & 2 \\
      \rowcolor{myLightBlue}\omit & Men et al. \cite{men2022gan} & Comput. Graph. 2022 & GAN & \cite{yun2012two, shu2016learning, shen2019interaction} & A & B & 2 \\
      & Goel et al. \cite{goel2022interaction} & Comput. Graph. Forum 2022 & GAN & \cite{yun2012two, shen2019interaction} & A & B & 2 \\
      \rowcolor{myLightBlue}\omit & Guo et al. \cite{guo2022multi} & CVPR 2022 & Transformer & \cite{guo2022multi} & N & B & 2 \\
      & DSAG \cite{gupta2023dsag} & WACV 2023 & VAE & \cite{liu2019ntu} & A & BH & 2+ \\
      \rowcolor{myLightBlue}\omit & InterFormer \cite{chopin2023interaction} & TOM 2023 & Transformer & \cite{kundu2020cross, yun2012two, hu2013efficient} & N & B & 2 \\
      & ActFormer \cite{xu2023actformer} & ICCV 2023 & GAN, Transformer & \cite{liu2019ntu, xu2023actformer} & A & B & 2+ \\
      \rowcolor{myLightBlue}\omit & RIG \cite{tanaka2023role} & ICCV 2023 & Diffusion & \cite{liu2019ntu} & A & B & 2 \\
      & MAMMOS \cite{lim2023mammos} & ICCV 2023 & VAE &  \cite{hassan2019resolving, replica19arXiv, AMASS:2019} & A & BH & 3+ \\
      \rowcolor{myLightBlue}\omit & BiGraphDiff \cite{chopin2024bipartite} & WACV 2024 & Diffusion & \cite{kundu2020cross, liu2019ntu} & A & B & 2 \\
      & InterGen \cite{liang2024intergen} & IJCV 2023 & Diffusion & \cite{liang2024intergen} & T & B & 2 \\
      \rowcolor{myLightBlue}\omit & ComMDM \cite{shafir2023human} & ICLR 2024 & Diffusion, Transformer & \cite{von2018recovering} & T & B & 2 \\
      & Duolando \cite{siyao2024duolando} & ICLR 2024 & VAE & \cite{siyao2024duolando} & M & BH & 2 \\      
      \rowcolor{myLightBlue}\omit & ReGenNet \cite{xu2024regennet} & CVPR 2024 & Diffusion & \cite{liang2024intergen, liu2019ntu, fieraru2020three} & A & BHF & 2 \\
      & Remos \cite{ghosh2024remos} & ECCV 2024 & Diffusion & \cite{shen2019interaction, guo2022multi, ghosh2024remos} & N & B & 2 \\
      \rowcolor{myLightBlue}\omit & FreeMotion \cite{fan2024freemotion} & ECCV 2024 & Framework & \cite{liang2024intergen} & T & B & number-free \\
      & Shan et al. \cite{shan2024towards} & ECCV 2024 & Diffusion & \cite{Guo_2022_CVPR, liang2024intergen, shan2024towards} & T & B & number-free \\
      \rowcolor{myLightBlue}\omit & InterControl \cite{wang2023intercontrol} & NeurIPS 2024 & Framework & \cite{mandery2015kit, Guo_2022_CVPR} & T & B & 2+ \\
      & TIM \cite{wang2024temporal} & arXiv 24.8 & RWKV & \cite{liang2024intergen} & T & B & 2 \\
      \rowcolor{myLightBlue}\omit & InterLDM \cite{li2024two} & arXiv 24.12 & VAE, Diffusion & \cite{liang2024intergen} & T & B & 2 \\
      & InterDance \cite{li2024interdance} & arXiv 24.12 & Diffusion & \cite{li2024interdance} & M & BH & 2 \\  
      \rowcolor{myLightBlue}\omit & TTR \cite{tanthink2025} & ICLR 2025 & VAE & \cite{xu2024inter, Guo_2022_CVPR} & A & B & 2 \\

      \addlinespace[-2pt]
      \bottomrule[0.4mm]
\label{methods}
\end{longtable}
\twocolumn

\subsection{Text-Conditioned Human-Scene Interaction}
In later works, action labels and textual descriptions are incorporated into human-scene interaction as semantic condition to enhance user control.
SAMP \cite{hassan2021stochastic} leverages GoalNet for predicting plausible action-dependent goal locations, a path planning module for obstacle-free navigation, and a CVAE-based MotionNet for diverse human-scene interactions. 
Wang et al. \cite{wang2022towards} propose a hierarchical framework utilizing CVAE for generating diverse scene-aware interaction anchors conditioned on action labels, a neural mapper with stochastic path planning for obstacle-free trajectories, and a Transformer-based motion completion network for generating motions between anchors.
HUMANISE \cite{wang2022humanise} introduces a CVAE-based generative model incorporating BERT \cite{devlin2019bert} for language understanding and a Transformer-based motion decoder to generate diverse and semantically consistent 3D human motions conditioned on both scene context and language descriptions.
MAMMOS \cite{lim2023mammos} employs a hierarchical framework that uses a CVAE for anchor placement tailored to human-scene and human-human interactions, a neural mapper with stochastic path generation for collision-free navigation, and a GRU-based CVAE for synthesizing smooth interaction motions, ensuring diverse and natural multi-human motion within complex 3D scenes.
Story-to-Motion \cite{qing2023story} deals with narrative-driven motion synthesis by employing a motion scheduler to structure long-text descriptions into motion segments, a text-based motion retrieval module to match motions with semantic and spatial constraints, and a progressive mask transformer to ensure smooth transitions, enabling infinite-length, controllable animations aligned with high-level storytelling.
Purposer \cite{ugrinovic2024purposer} employs a discrete autoregressive model with a two-branch architecture for integrating scene geometry, semantic goals, and past-future motion contexts, mapping interaction motion into a discrete latent space.
Cen et al. \cite{cen2024generating} leverages large language model for 3D object localization and diffusion model for trajectory and motion synthesis, enabling semantically consistent human-scene interactions based on text descriptions.
Wang et al. \cite{wang2024move} propose a two-stage diffusion-based framework that employs an affordance diffusion model for generating language-grounded affordance maps and an affordance-to-motion diffusion model for synthesizing plausible human motions.
TesMo \cite{yi2024generating} similarly employs a two-stage pipeline to generate goal-directed locomotion while avoiding obstacles and synthesizing object-specific human motions from textual descriptions.
SCENIC \cite{zhang2024scenic} addresses interaction generation on complex terrains by employing hierarchical scene reasoning for fine-grained adaptation, goal-centric canonicalization for high-level navigation planning, and frame-wise text alignment for precise semantic control, enabling physically plausible and text-conditioned human interaction motion synthesis in diverse 3D environments.

\subsection{Extensive Human-Scene Interaction}
Recent human-scene interaction approaches have been extended to encompass human-human and human-object interactions, enabling more comprehensive interactions within both static and dynamic scenes.
TRUMANS \cite{jiang2024scaling} tackles both HSI and HOI by employing a local scene perceiver for collision avoidance and environment adaptation in HSI, while utilizing an action-conditioned diffusion module to progressively generate HOI motion through episodic segment.
LINGO \cite{jiang2024autonomous} presents an diffusion-based framework that unifies human-scene interaction and human-object interaction into a seamless pipeline, leveraging a dual voxel scene encoder and a joint time frame-language embedding to ensure contextually accurate motion generation in complex 3D environments.
LaserHuman \cite{cong2024laserhuman} proposes a multi-conditional diffusion framework for human scene and human-human interaction generation, utilizing a fusion module to integrate textual and 3D scene constraints, enabling semantically consistent and physically plausible motion synthesis in both indoor and outdoor free environments.
ZeroHSI \cite{li2024zerohsi} introduces a zero-shot interaction motion generation framework by distilling motion priors from video generation models and using differentiable neural rendering to reconstruct realistic human motions, enabling human-scene and human-object interactions in both static and dynamic 3D environments without requiring paired motion capture data.
Sitcom-Crafter \cite{chen2024sitcom} introduces a modular motion generation system that unifies human-scene and human-human interaction, leveraging an augmentation modules for motion synchronization and collision avoidance, producing high-quality character animation guided by plot-driven narratives in 3D scenes.

\subsection{Human-Scene Interaction Prediction}
Unlike general human-scene interaction generation, which optionally takes a starting pose as an input condition, human-scene interaction prediction focuses on producing a short sequence of interaction motion based on history motions of similar duration.
Cao et al. \cite{cao2020long} propose a three-stage framework using CVAE for goal prediction, a heatmap-based network for path planning, and a Transformer for pose refinement to generate scene-aware 3D human motions that consider both spatial and contextual constraints.
Mao et al. \cite{mao2022contact} introduce a two-stage pipeline based on GRU and CNN, leveraging distance-based contact maps to predict future human-scene interactions by explicitly modeling human-scene contacts.
DiMoP3D \cite{lou2024harmonizing} introduces a diffusion-based motion prediction framework that harmonizes stochastic human motion with deterministic 3D scene constraints, leveraging a context-aware intermodal interpreter for human intention estimation, a stochastic planner for trajectory and end-pose prediction, and a self-prompted motion generator for diverse and physically consistent human-scene interactions.

\section{Human-Object Interaction} \label{Human-Object Interaction}
The human-object interaction generation task focuses on the synchronized generation of human motion and the corresponding motion of interacting objects. These tasks can be considered a specialized subset of human-scene interaction, as static or dynamic objects can be regarded as components of the scene. However, human-object interaction specifically emphasizes the accurate execution of approaching, grasping, and manipulating the objects.

\subsection{Whole-Body Hand Object Interaction}
Early human-object interaction generation works focus on whole-body hand object interaction motion generation. Derived from hand grasp generation task, these works extend hand grasping into whole-body range hand object interaction. Though only hands directly participate in the interaction, other body parts coordinately move together when human reaches out and manipulates for the objects.
SAGA \cite{wu2022saga} employs a multi-task CVAE to generate the final grasping pose and then infills the motion between the start and end poses with a contact-aware module to model whole-body hand-object interactions.
Similarly, GOAL \cite{taheri2022goal} leverages GNet and MNet to generate a static whole-body grasping pose and then autoregressively infills the motion sequence towards the grasp, achieving realistic coordination of facial, body, and hand movements.
Later works extend the interaction from approaching to further object manipulation, synchronously generate both human motion and corresponding object motion.
Razali et al. \cite{razali2023action} generate action-conditioned bimanual object manipulation sequences, where object motion is predicted using a graph recurrent network, finger poses are estimated with an attention-based recurrent network, and full-body poses are reconstructed to ensure realistic interaction synthesis.
IMoS \cite{ghosh2023imos} generates whole-body hand-object interactions using a pair of decoupled CVAEs for the arms and body, optimized with an object position module to ensure plausible object poses. Moreover, it employs action-object pairs as semantic intentions to guide the generation.
OMOMO \cite{li2023object} leverages conditional diffusion models to generate high quality human manipulation motions guided by object movements, first predicting hand positions and applying contact constraints, then generating whole-body poses based on the corrected hand positions, ensuring physically plausible interactions with objects.
TOHO \cite{li2024task} generates continuous task-oriented human-object interaction motions by leveraging implicit neural representations, first predicting keyframe poses, then using a motion infilling network for smooth transitions, and finally applying a closed-form object motion estimation algorithm to ensure consistent object movements.
OOD-HOI \cite{zhang2024ood} focuses on out-of-domain hand-object interaction generation by employing a dual-branch reciprocal diffusion model for initial pose generation and a dynamic adaptation mechanism that incorporates semantic adjustment and geometry deformation to enhance robustness against unseen actions and objects.
Diffgrasp \cite{zhang2024diffgrasp} generates whole-body human grasping sequences guided by object motion by jointly modeling body, hands, and object interactions, ensuring physically plausible and dynamically coordinated grasps.

\subsection{Human-Object Interaction Generation}
Human-object interaction involving diverse body parts as well evolves from static object interaction to dynamic objects.
COUCH \cite{zhang2022couch} proposes a method for generating contact-based human-chair interactions by utilizing ControlNet to predict hand trajectories towards specified contact points and PoseNet to synthesize whole-body motion.
Pi et al. \cite{pi2023hierarchical} also propose a hierarchical generation framework by generating milestones points and poses along the motion trajectory and infill motions between milestones.
Other than human-chair interactions, conventional human-object interaction includes all types of dynamic objects.
InterDiff \cite{xu2023interdiff} generates human-object interaction sequences with a diffusion network and employs a physics-aware interaction correction module leveraging relative motion patterns around contact points to ensure physical plausibility.
To generate controllable human-object interactions driven by natural language descriptions, later works encode text using pretrained CLIP \cite{radford2021learning} or LLMs \cite{achiam2023gpt, touvron2023llama} to better capture semantic meaning.
ROAM \cite{zhang2024roam} generates object-aware human interaction motions that generalize to unseen objects using an SE(3)-equivariant neural descriptor field to optimize goal poses from a single reference object, combined with a Neural State Machine for seamless motion synthesis, enabling robust character interactions without requiring large motion capture datasets.
NIFTY \cite{kulkarni2024nifty} employs an object-centric interaction field to guide a diffusion-based motion generator, producing diverse and realistic last-mile interactions.
HOI-Diff \cite{peng2023hoi} uses a dual-branch diffusion model for coarse motion generation and an affordance prediction diffusion model to estimate contact areas, ensuring physically plausible and semantically aligned interactions across diverse objects.
HOIAnimator \cite{song2024hoianimator} generates text-driven human-object interactions using a dual perceptive diffusion model that separately models human and object movements while ensuring interaction consistency through a perceptive message passing mechanism.
CG-HOI \cite{diller2024cg} jointly models human motion, object motion, and contact using cross-attention in a diffusion-based structure and employs contact-based guidance during inference to generate physically plausible interactions.
CHOIS \cite{li2025controllable} generates synchronized human-object interactions guided by text descriptions and sparse object waypoints, using a conditional diffusion model enhanced by geometry loss and contact guidances to ensure semantically aligned and physically plausible interactions.
THOR \cite{wu2024thor} enhances dynamic human-object interaction motions by a human-object relation intervention mechanism, which refines object motion through kinematic and geometric relations.

Recent works diverse in innovative emphasis on interaction characteristics.
HOI-M3 \cite{zhang2024hoi} focuses on multi-human and multi-object interactions in contextual environments, offering a baseline method for multi-HOI generation, facilitating complex interaction modeling in social settings.
HIMO-Gen \cite{lv2024himo} proposes a dual-branch conditional diffusion model with a mutual interaction module for generating human interaction with multiple objects, along with an autoregressive pipeline for smooth multi-step interaction generation.
InterDreamer \cite{xu2024interdreamer} generates zero-shot human-object interaction sequences guided by text by decoupling semantics and dynamics. It leverages a LLM for high-level planning, a text-to-motion model for initial human motion generation, and a world model for physics-informed object dynamics, achieving semantically aligned interactions without relying on text-interaction paired data.
Wu et al. \cite{wu2024human} uses a two-stage framework with an LLM-based high-level planner that generates task plans and a low-level motion generator that sequentially synthesizes synchronized full-body, object, and finger motions.
SyncDiff \cite{he2024syncdiff} synthesizes multiple humans and objects interaction using a single diffusion model that jointly captures individual and relative motions, introducing a frequency-domain motion decomposition for fine-grained motion fidelity.
TriDi \cite{petrov2024tridi} introduces a unified trilateral diffusion model for generating human-object interactions by jointly modeling humans, objects, and interactions as a three-variable distribution, utilizing tokenized representations and a shared latent space that enables flexible, realistic generation across seven operation modes.
FORCE \cite{zhang2024force} leverages a kinematic-based method for generating nuanced human-object interaction motions by leveraging an intuitive physics encoding that models the relation of human force and object resistance to enable physically informed interaction synthesis.
AvatarGO \cite{cao2024avatargo} as well generates zero-shot human-object interaction guided by textual descriptions using a framework that integrates LLM-guided contact retargeting and correspondence-aware motion optimization for coherent animation.

\subsection{Human-Object Interaction Prediction}
A specific type of human-object interaction generation is human-object interaction prediction. Similar to human-scene interaction generation, it aims to forecast future interactions over a temporal duration comparable to the input history poses.
CATMO \cite{corona2020context} generates contact-aware interaction motion sequences from textual descriptions by independently encode motion and contact with VQ-VAEs and then autoregressively generates with intertwined GPT.
HO-GCN \cite{wan2022learn} leverages a graph convolutional network to predict human-object interaction motions by integrating an object dynamic descriptor that encodes intrinsic physical properties with spatial-temporal graph convolution to achieve generalizable motion prediction.
HOIMotion \cite{hu2024hoimotion} forecasts human motion by combining past body poses and egocentric 3D object bounding boxes through a graph-based encoder-residual-decoder architecture.

\section{Human-Human Interaction} \label{Human-Human Interaction}

Humans frequently interact with one another, either through actions initiated by an active participant or through mutual involvement in the interaction. This section introduces the task of human-human interaction generation, which can be broadly categorized into three main types: human interaction reaction, text-guided multi-human interaction, and human-human interaction prediction.

\subsection{Human Interaction Reaction}
Human interaction reaction generation focuses on modeling interactions between an actor and a reactor, where the goal is to generate the reactor's motion in response to a given active motion initiated by the actor.
Men et al. \cite{men2022gan} first propose a semi-supervised GAN-based model for human interaction reaction generation, leveraging part-based long short-term memory networks and attention mechanisms to capture spatiotemporal features, while introducing a class-aware discriminator to generate class-specific motions.
Goel et al. \cite{goel2022interaction} further leverage a conditional hierarchical GAN to generate controllable two-character close interactions, producing spatiotemporally aligned motions by enabling interaction mix-and-match capabilities.
InterFormer \cite{chopin2023interaction} generates complex and long-term reactive motions using a Transformer-based architecture that integrates spatial and temporal attention modules to model both character motion and their interactions.
BiGraphDiff \cite{chopin2024bipartite} uses bipartite graphs to model interactions between skeleton nodes and a Transformer-based diffusion architecture to capture spatiotemporal dependencies.
ReGenNet \cite{xu2024regennet} introduces an explicit distance-based interaction loss to capture asymmetric and synchronous human interactions, enabling instant and plausible online reaction generation even under unseen motions and viewpoint changes.
Remos \cite{ghosh2024remos} proposes a diffusion model with spatio-temporal cross-attention and hand-interaction-aware attention mechanisms, enabling fine-grained and synchronized full-body and hand motion generation.
TTR \cite{tanthink2025} introduces a two-stage framework where the model first infers action intent through a reasoning step and then predicts semantically appropriate reactions, while leveraging a unified motion tokenizer that decouples egocentric pose and absolute space features to enhance motion representation and generalization.
Duolando \cite{siyao2024duolando} proposes a GPT-based model for dance reaction generation, leveraging a large-scale duet dance dataset and an interaction-coordinated GPT to enhance synchronization and responsiveness in two-person dance interactions.
InterDance \cite{li2024interdance} as well generates reactive human dance motion with an interaction refinement guidance strategy to progressively optimize realism, ensuring accurate hand-body contact and penetration-free interactions.

\subsection{Text-guided Multi-Human Interaction}
Text-guided multi-human interaction generation aims to generate corresponding multi-human interaction motions from free-form natural language descriptions.
MUGL \cite{maheshwari2022mugl} uses a Conditional Gaussian Mixture VAE and hybrid pose representation to generating multi-person interaction motion sequences with locomotion.
DSAG \cite{gupta2023dsag} generates action-conditioned multi-actor whole-body interaction, introducing dedicated finger joint representations, spatiotemporal transformation blocks with multi-head self-attention, and specialized temporal processing to enhance motion realism.
ActFormer \cite{xu2023actformer} leverages a Transformer architecture under a GAN training scheme for action-conditioned human reaction generation, integrating a Gaussian Process latent prior to enhance temporal correlations and spatial dynamics.
RIG \cite{tanaka2023role} proposes a role-aware interaction generation model that enables asymmetric interactions by designating active and passive roles based on textual descriptions, leveraging dual Transformers with shared parameters and cross-attention modules to achieve mutual motion consistency.
MAMMOS \cite{lim2023mammos} generates multi-human interactions in large-scale 3D scenes, introducing eye-contact optimization and interaction-aware modules to ensure physical plausibility and realistic human-human interactions.
InterGen \cite{liang2024intergen} introduces cooperative Transformer-based denoisers with shared weights and mutual attention to model two-person symmetrical interactions.
ComMDM \cite{shafir2023human} leverages pretrained Motion Diffusion Models \cite{tevet2023human} and a lightweight communication module to address two-person interaction, enabling few-shot multi-person interaction motion generation.
FreeMotion \cite{fan2024freemotion} proposes a unified framework for number-free multi-human interaction with a recursive conditional motion modeling paradigm that supports any number of participants.
Shan et al. \cite{shan2024towards} also propose an open-domain framework capable of generating multi-human motion sequences for an arbitrary number of subjects.
InterControl \cite{wang2023intercontrol} enables precise spatial control of every joint with a Motion ControlNet for adaptive joint alignment and an inverse kinematics loss for fine-grained joint positioning.
TIM \cite{wang2024temporal} leverages Receptance Weighted Key Value for efficient human-human interaction motion generation, significantly reducing model parameters.
InterLDM \cite{li2024two} encodes two-person motions into a unified latent space and efficiently generates asymmetrical and context-consistent interactions with a conditional latent diffusion model.

\subsection{Human-Human Interaction Prediction}
Human-human interaction prediction refers to the task of modeling and forecasting the dynamic interactions between multiple humans based on historical motion data.
ccRNN \cite{kundu2020cross} proposes a cross-conditioned recurrent network for long-term prediction of human-human interactions, integrating autoregressive and encoder-decoder architectures in a hierarchical manner to enable synchronized motion generation for extended durations.
Baruah et al. \cite{baruah2020multimodal} leverages an unsupervised attention-based agent that actively selects and attends to informative body regions using saliency maps derived from prediction errors, achieving efficient human-human interaction prediction.
Guo et al. \cite{guo2022multi} predict multi-person extreme motion with a cross-interaction attention module, introducing a collaborative motion forecasting approach that exploits historical pose dependencies between interacting individuals to improve motion realism in highly dynamic interactions.

\section{Datasets and Evaluation Metrics} \label{Dataset and Evaluation Metrics}
In this section, we summarize the datasets and evaluation metrics for human interaction generation. In Table \ref{datasets}, we show the collection method, the representations, the statistics data, and the contact information of these datasets.

\subsection{Human-Scene Interaction Datasets}
Human-scene interaction generation datasets primarily include human interaction motion data, static or dynamic 3D scene models, and semantic annotations that capture diverse and context-aware human-scene interactions.

\textbf{3DPW} \cite{von2018recovering} contains 60 sequences with over 51K frames of real world interactions, recorded using a single handheld camera synchronized with IMUs. It includes 3D poses and shapes that are annotated with SMPL body models.

\textbf{PROX} \cite{hassan2019resolving} includes 0.1M synchronized RGB-D frames across 12 indoor scenes, featuring 20 subjects performing natural interactions. Motion data is recorded using Kinect-One cameras and Structure Sensor, providing SMPL-X models with pseudo-ground truth for body poses, as well as annotations of contact and penetration constraints between the human and the scene.

\textbf{GTA-IM} \cite{cao2020long} provides human motion captured across 10 diverse indoor scenes featuring 50 characters performing activities like walking, sitting, and interacting with objects. Data was synthetically generated using the GTA engine, incorporating randomized goals, tasks, and viewing angles for diverse human-scene interaction scenarios.

\textbf{HPS} \cite{guzov2021human} collects 300K IMU data from 7 subjects across 8 large indoor and outdoor scenes. The dataset provides human poses registered with prescanned 3D scene reconstructions. Motion data was captured using a head-mounted camera and body-mounted IMUs, enabling drift-free human motion estimation and self-localization in large-scale environments.

\textbf{SAMP} \cite{hassan2021stochastic} includes 185K motion capture frames across 7 object categories, featuring dynamic human-scene interactions. Motion data is captured using Vicon cameras, providing SMPL-X body models with contact points and CAD object models. The dataset includes augmented interactions with 13K synthetic samples generated using ShapeNet \cite{chang2015shapenet} objects to enhance diversity and generalization to unseen geometries.

\textbf{HSC4D} \cite{dai2022hsc4d} collects 250K mocap frames across 3 large-scale indoor and outdoor scenes. It provides SMPL human body models, LiDAR point clouds, and optimized global trajectories for human motion and environment mapping. The interaction data was captured using body-mounted IMUs and a hip-mounted LiDAR system, enabling robust human-scene interaction modeling in diverse settings.

\textbf{RICH} \cite{huang2022capturing} includes 142 interaction motion sequences captured across 5 indoor and outdoor static scenes, featuring 22 subjects interacting with their environments. The dataset provides SMPL-X human body models with dense vertex-level human-scene contact labels and high-quality 3D scene scans. Data was recorded using synchronized static cameras and laser scanning for scene reconstruction, enabling detailed 3D contact estimation and human-scene interaction analysis.

\textbf{GIMO} \cite{zheng2022gimo} includes 129K IMU-based motion capture data and 3D eye gaze annotations across 19 indoor scenes. It features 11 subjects performing 217 motion trajectories with clear semantic intentions, providing SMPL-X models, 3D scene reconstructions, and gaze-informed motion predictions. The dataset was collected using head-mounted AR devices and LiDAR-equipped smartphones, enabling diverse and dynamic human-scene interaction modeling.

\textbf{HUMANISE} \cite{wang2022humanise} captures 19.6K high-quality human motion sequences across 643 3D indoor scenes, annotated with 1.2M frames and template-based language descriptions. Data is synthetically generated by aligning AMASS \cite{AMASS:2019} motions with ScanNet \cite{dai2017scannet} 3D scene reconstructions, providing SMPL-X body models, interaction annotations, and semantic action labels.

\textbf{CIRCLE} \cite{araujo2023circle} consists of 10 hours of motion capture reaching sequences recorded from 5 subjects performing tasks across 9 virtual indoor household scenes. Motion data was collected using a Vicon optical motion capture system synchronized with a Meta Quest 2 VR headset running Habitat \cite{savva2019habitat}, enabling scene-aware human motion capture in highly contextual environments.

\textbf{CIMI4D} \cite{yan2023cimi4d} contains 180 minutes of IMU data collected from 12 subjects climbing 13 walls across indoor and outdoor settings. The dataset provides SMPL body models, high-precision static and dynamic scene point clouds, annotated human-rock contact points, and global trajectories, enabling detailed analysis of human-scene interactions during complex climbing motions.

\textbf{BEDLAM} \cite{black2023bedlam} provides SMPL-X body models with physics-simulated clothing, hair, and high-resolution textures, captured across 95 HDRI environments and 8 3D scenes with up to 10 people per scene. The dataset includes depth maps, segmentation masks, and multi-view sequences, enabling 3D human-scene interaction generation from synthetic data.

\textbf{SLOPER4D} \cite{dai2023sloper4d} includes 500K IMU-based human motion frames collected from 12 subjects across 10 large-scale urban environments. The dataset provides global 3D human poses, reconstructed 3D scene point clouds, and multi-view annotations for 3D human keypoints.

\textbf{EgoHOI} \cite{guzov2024interaction} collects 162K IMU-based motion frames and 54K egocentric video frames across 7 interactive indoor environments, featuring human interactions such as opening doors, moving furniture, and adjusting scene elements. It provides SMPL body models, contact annotations, enabling full human-scene interaction modeling.

\textbf{ScenePlan} \cite{xiao2024unified} consists of 1K task plans generated from diverse indoor scenarios, with thousands of interaction steps represented as Chains of Contacts (CoC). The dataset spans various interaction complexities and involves multiple objects across PartNet \cite{mo2019partnet} and ScanNet \cite{dai2017scannet} scenes. It provides language-based task plans, object-part annotations, and realistic motion data generated using large language models for task planning.

\textbf{TRUMANS} \cite{jiang2024scaling} includes 1.6M frames of motion capture data across 100 indoor scenes, featuring whole-body human motions interacting with 20 object categories, including static and dynamic objects. The dataset also provides SMPL-X body models, per-vertex contact annotations, multi-view and ego-centric RGBD videos, and augmented interactions using virtual models for diverse scenes and actions.

\textbf{RELI11D} \cite{yan2024reli11d} contains 239K frames of multimodal human motion data, captured across 7 different sports performed by 10 actors in 5 diverse sports scenes. The dataset integrates RGB video, LiDAR point clouds, and IMU motion capture data, offering 3.32 hours of synchronized data. It provides high-precision 3D human pose annotations using SMPL models, global motion trajectories, and detailed 3D scene reconstructions.

\textbf{LINGO} \cite{jiang2024autonomous} includes 16 hours of mocap sequences across 120 indoor scenes, featuring 40 motion types such as locomotion, grasping, and human-scene interactions, all annotated with precise language descriptions. The dataset provides detailed motion trajectories and scene representations, enabling text-guided human-scene and human-object interaction generation.

\textbf{LaserHuman} \cite{cong2024laserhuman} collects 3K human motion sequences with 12K language descriptions in 11 large-scale indoor and outdoor scenes. It provides SMPL body models, dynamic LiDAR point clouds, 3D scene reconstructions, and fine-grained text descriptions for human-scene interactions, enabling language-guided human-scene interaction generation in complex and dynamic environments.

\textbf{SCENIC} \cite{zhang2024scenic} includes 15K sequences of human motion adapted to diverse terrains with annotated text prompts. The dataset provides SMPL body models, terrain-aware human motion trajectories, and hierarchical scene reasoning representations, enabling scene-aware motion synthesis in complex 3D environments.

\subsection{Human-Object Interaction Datasets}
Human-object interaction generation datasets basically includes human motion and corresponding object motion data. Beside the generation task, available datasets also come from human-object estimation and motion capture field.

\textbf{KIT} \cite{mandery2015kit} contains motion sequences of 43 participants interacting with 41 objects, captured using a Vicon MX motion capture system. The dataset also provides hierarchical motion description labels to facilitate structured searches and categorization. Human motion data is normalized using the Master Motor Map (MMM) framework.

\textbf{PiGraphs} \cite{savva2016pigraphs} includes 63 motion capture human interaction observations across 30 reconstructed indoor 3D scenes, featuring 0.1M frames with 298 annotated actions spanning 43 verb-noun pairs. Motion data is recorded using Kinect v2 RGB-D sensors, providing 3D skeletal joint positions, voxel-based object annotations, and interaction graphs encoding human pose and object geometry relationships.

\textbf{GRAB} \cite{taheri2020grab} collects whole-body human motion sequences of 10 participants manipulating 51 everyday objects, captured using Vicon infrared cameras. It includes contact heatmaps and four categories of action labels to enrich the data annotations. Human motion is parameterized using the SMPL-X model, integrating body, face, and hands, while the objects are represented as 3D meshes.

\textbf{GraviCap} \cite{dabral2021gravity} captures 9 sequences human-object interactions involving up to two participants and various spherical objects undergoing free flights, recorded using monocular RGB cameras in multiple-view setups. It includes SMPL-based human motion data, 3D object trajectories computed under gravity constraints, and contact joints annotations.

\textbf{D3D-HOI} \cite{xu2021d3d} comprises interactions between humans and articulated objects, including 5 participants and 8 object categories. The dataset also provides object part motion based on articulated CAD models derived from the PartNet-Mobility dataset \cite{xiang2020sapien}. The dataset includes 6,286 annotated frames, offering contextual interactions for articulated 3D object interaction.

\textbf{Human-Large Object Interaction Dataset (HLOI)} \cite{wan2022learn} includes motion sequences of 6 participants interacting with 12 large-sized objects, captured using an OptiTrack motion capture system. It provides 3D skeletal human motion data, 6DoF object pose data represented by 12 keypoints per object, and annotated action labels.

\textbf{BEHAVE} \cite{bhatnagar2022behave} contains 15.2K frames of 8 participants interacting with 20 objects, captured across 5 indoor locations using a portable multi-camera RGBD setup. It includes annotated human-object contact vertices, object segmentation masks, and pseudo-ground truth 3D human and object fits to enhance interaction analysis.

\textbf{InterCap} \cite{huang2022intercap} captures 67K multi-view frames of 10 participants interacting with 10 objects across 223 sequences using Azure Kinect RGB-D cameras. It includes RGB-D videos and contact heatmap derived from mesh proximity metrics. Human motion is represented using SMPL-X and objects are represented as prescanned 3D meshes.

\textbf{COUCH} \cite{zhang2022couch} provides human motion sequences of 6 participants interacting with chairs or sofas, captured in four different indoor scenes using Kinect sensors and IMUs. Human motion is parameterized using the SMPL model, while the chairs are represented as meshes. The dataset features over 500 motion sequences spanning a total of 3 hours of interaction.

\textbf{HoDome} \cite{zhang2023neuraldome} captures detailed human-object interactions involving 10 participants engaging with 23 diverse objects across 274 sequences, recorded using RGB cameras and OptiTrack motion capture cameras. The dataset provides pseudo contact labels derived from proximity thresholds. Human motion is parameterized using the SMPL-X model for body, face, and hand dynamics, while objects are represented as prescanned 3D meshes.

\textbf{CHAIRS} \cite{jiang2022chairs} includes whole-body human motion sequences involving 46 participants interacting with 81 articulated and rigid sitable objects, captured using a hybrid motion capture system combining Azure Kinect RGB-D cameras and inertial-optical tracking systems. The objects are represented as prescanned and refined 3D meshes, segmented into functional parts.

\textbf{FullBodyManipulation} \cite{li2023object} consists of 10 hours of paired human motion and object motion data for 17 participants interacting with 15 large-sized objects, captured using a Vicon motion capture system. Additionally, object motions are captured using an iPhone with ARKit, and the object is represented with a Basis Point Set.

\textbf{CoChair} \cite{liu2023interactive} captures large-scale human-object-human interactions, featuring 10 pairs of participants collaboratively manipulating 8 different chairs, recorded using NOKOV motion capture cameras. The dataset includes SMPL-X human motion data and 6D CAD object poses, enabling the study of multi-human-object interaction generation.

\textbf{$\text{HOI-M}^\textbf{3}$} \cite{zhang2024hoi} consists of multi-human and multi-object interaction data involving 31 participants engaging with 90 diverse objects, recorded using synchronized 4K cameras and object-mounted IMUs. Human motion is parameterized using the SMPL model and objects are represented as prescanned 3D meshes.

\textbf{$\text{IMHD}^\textbf{2}$} \cite{zhao2024m} captures human-object interactions involving 15 participants interacting with 10 objects across 295 sequences, using synchronized 4K RGB cameras and object-mounted IMUs. It includes human motion data represented by SMPL-H model, object motion with 6D poses derived from IMUs, and high-quality 3D object scans.

\textbf{HIMO} \cite{lv2024himo} collects human motion interacting with multiple objects, featuring 34 participants engaging with 53 diverse household objects across 3K sequences. Motion data is recorded using a hybrid motion capture system with OptiTrack infrared cameras and Noitom Perception Neuron Studio gloves. The dataset also includes fine-grained textual descriptions with temporal segmentation for detailed interaction modeling.

\textbf{ParaHome} \cite{kim2024parahome} captures human-object interactions in a room-scale environment, featuring 30 participants engaging with 22 objects across 101 scenarios. Motion data is recorded using synchronized RGB cameras, Xsens IMU suits, and Manus gloves. The dataset includes SMPL-X motion data for body and hand dynamics, 3D mesh objects with articulation parameters, and textual instructions.

\textbf{CORE4D} \cite{zhang2024core4d} captures collaborative human-object-human interactions, featuring 37 objects from 6 categories and over 1K real-world sequences. Additionally, synthetic data extends the dataset to 10K sequences with 3K diverse object geometries. Spanning diverse collaboration modes and dynamic 3D scenes, CORE4D provides a large-scale resource for studying multi-person and object interactions with real and synthetic components.

\textbf{FORCE} \cite{zhang2024force} includes human motion sequences involving interactions with 8 objects under varying resistance levels, captured using Kinect RGB-D cameras and IMU sensors. It provides annotations of physical properties, including resistance, force, and contact modes.

\renewcommand\arraystretch{1.3}
\setlength\tabcolsep{2pt}
\setlength{\LTcapwidth}{\textwidth}
\setlength{\LTleft}{0pt}
\setlength{\LTright}{0pt}

\scriptsize
\onecolumn
\begin{longtable}{
    >{\centering\arraybackslash}m{0.03\textwidth}
    >{\centering\arraybackslash}m{0.17\textwidth}
    >{\centering\arraybackslash}m{0.16\textwidth}
    >{\centering\arraybackslash}m{0.10\textwidth}
    >{\centering\arraybackslash}m{0.12\textwidth}
    >{\centering\arraybackslash}m{0.13\textwidth}
    >{\centering\arraybackslash}m{0.06\textwidth}  
    >{\centering\arraybackslash}m{0.06\textwidth}
    >{\centering\arraybackslash}m{0.10\textwidth}
}
    \caption{Datasets for human interaction generation. "H Rep", "E Rep", "Sub", "Ent" represent Human Representation, Entity Representation, Subjects, and Entities, respectively. In human-human interaction datasets, "E Rep", "Ent", and "Contacts" denote asymmetry annotation, hand motion, and text annotation, respectively.} \\
    \toprule[0.4mm]
    \addlinespace[0.2mm]
    \cmidrule[0.6pt]{1-9}
    & \textbf{Dataset} & \textbf{Venue} & \textbf{Collection} & \textbf{H Rep} & \textbf{E Rep} & \textbf{Sub} & \textbf{Ent} & \textbf{Contact}\\
    \endfirsthead
    
    \toprule[0.4mm]
    \addlinespace[0.2mm]
    \cmidrule[0.6pt]{1-9}
    & \textbf{Dataset} & \textbf{Venue} & \textbf{Collection} & \textbf{H Rep} & \textbf{E Rep} & \textbf{Sub} & \textbf{Ent} & \textbf{Contact}\\
    \midrule
    \endhead

    \midrule
    \multicolumn{9}{r}{\vspace{-5pt} \raisebox{4pt}{\textit{Continued on next page}}} \\ 
    \midrule
    \endfoot
    \endlastfoot

      \addlinespace[-1.0pt]
      \midrule
      \addlinespace[1.0pt]
      \multirow{20}{*}{\makecell{\rotatebox{90}{\textbf{HSI Datasets}}}}
      & 3DPW \cite{von2018recovering} & ECCV 2018 & Inertial-Based & SMPL & RGB Videos & 7 & - & None \\
      \noalign{\global\renewcommand{\arraystretch}{1.3}}
      
        \rowcolor{myLightYellow}\omit & PROX \cite{hassan2019resolving} & ICCV 2019 & Markerless & SMPL-X & Mesh & 20 & 12 & Contact Vertex \\
	& GTA-IM \cite{cao2020long} & ECCV 2020 & Synthetic & 3D Skeleton & RGB Images & 50 & 49 & None \\
	\rowcolor{myLightYellow}\omit & HPS \cite{guzov2021human} & CVPR 2021 & Inertial-Based & SMPL & Point Cloud & 7 & 8 & Foot Contact \\
	& SAMP \cite{hassan2021stochastic} & ICCV 2021 & Marker-Based & SMPL-X & Voxel & - & - & Contact Part \\
	\rowcolor{myLightYellow}\omit & HSC4D \cite{dai2022hsc4d} & CVPR 2022 & Inertial-Based & SMPL & Point Cloud & - & 3 & None \\
	& RICH \cite{huang2022capturing} & CVPR 2022 & Markerless & SMPL-X & Mesh & 22 & 5 & Contact Vertex \\
	\rowcolor{myLightYellow}\omit & GIMO \cite{zheng2022gimo} & ECCV 2022 & Inertial-Based & SMPL-X & Mesh & 11 & 19 & None \\
	& HUMANISE \cite{wang2022humanise} & NeurIPS 2022 & Synthetic & SMPL-X & Point Cloud & - & 643 & None \\
	\rowcolor{myLightYellow}\omit & CIRCLE \cite{araujo2023circle} & CVPR 2023 & Marker-Based & SMPL-X & Point Cloud & 5 & 9 & None \\
	& CIMI4D \cite{yan2023cimi4d} & CVPR 2023 & Inertial-Based & SMPL & Point Cloud & 12 & 13 & Contact Part \\
	\rowcolor{myLightYellow}\omit & BEDLAM \cite{black2023bedlam} & CVPR 2023 & Synthetic & SMPL-X & HDRI Images & 271 & 95+8 & None \\
	& SLOPER4D \cite{dai2023sloper4d} & CVPR 2023 & Inertial-Based & SMPL & Point Cloud & 12 & 10 & None \\
	\rowcolor{myLightYellow}\omit & EgoHOI \cite{guzov2024interaction} & 3DV 2024 & Inertial-Based & SMPL & Point Cloud & 4 & 7 & Contact Part \\   
	& ScenePlan \cite{xiao2024unified} & ICLR 2024 & Synthetic & SMPL-X & Point Cloud & - & 10 & CoC \\
	\rowcolor{myLightYellow}\omit & TRUMANS \cite{jiang2024scaling} & CVPR 2024 & Marker-Based & SMPL-X & Voxel & 7 & 100 & Contact Vertex \\
	& RELI11D \cite{yan2024reli11d} & CVPR 2024 & Inertial-Based & SMPL & Point Cloud & 10 & 7 & None \\
	\rowcolor{myLightYellow}\omit & LINGO \cite{jiang2024autonomous} & SIGGRAPH Asia 2024 & Marker-Based & SMPL-X & Voxel & - & 120 & None \\
	& LaserHuman \cite{cong2024laserhuman} & arXiv 24.3 & Hybrid & SMPL & Point Cloud & - & 11 & None \\
	\rowcolor{myLightYellow}\omit & SCENIC \cite{zhang2024scenic} & arXiv 24.12 & Synthetic & SMPL & Distance Field & - & - & None \\

      \addlinespace[-1.0pt]
      \midrule
      \addlinespace[1.0pt]
      \multirow{19}{*}{\makecell{\rotatebox{90}{\textbf{HOI Datasets}}}}
      & KIT \cite{mandery2015kit} & ICAR 2015 & Marker-Based & MMM & Rigid Body & 43 & 41 & None \\
      \noalign{\global\renewcommand{\arraystretch}{1.3}}

        \rowcolor{myLightYellow}\omit & PIGraphs \cite{savva2016pigraphs} & TOG 2016 & Markerless & 3D Skeleton & - & 5 & 19 & Contact Joint \\
        & GRAB \cite{taheri2020grab} & ECCV 2020 & Marker-Based & SMPL-X & Mesh & 10 & 51 & Heatmap \\
        \rowcolor{myLightYellow}\omit & GraviCap \cite{dabral2021gravity} & ICCV 2021 & Synthetic & 3D Skeleton & Geometric Center & 4 & 4 & Contact Joint \\
        & D3D-HOI \cite{xu2021d3d} & arXiv 21.8 & Synthetic & SMPL & CAD & 5 & 24 & None \\
        \rowcolor{myLightYellow}\omit & HLOI \cite{wan2022learn} & ICRA 2022 & Marker-Based & 3D Skeleton & Keypoints & 6 & 12 & None \\
        & BEHAVE \cite{bhatnagar2022behave} & CVPR 2022 & Markerless & SMPL & Mesh & 8 & 20 & Contact Vertex \\
        \rowcolor{myLightYellow}\omit & InterCap \cite{huang2022intercap} & GCPR 2022 & Synthetic & SMPL-X & Mesh & 10 & 10 & Heatmap \\
        & COUCH \cite{zhang2022couch} & ECCV 2022 & Hybrid & SMPL & Mesh & 6 & - & Contact Part \\
        \rowcolor{myLightYellow}\omit & HODome \cite{zhang2023neuraldome} & CVPR 2023 & Hybrid & SMPL-X & Mesh & 10 & 23 & Contact Part \\
        & CHAIRS \cite{jiang2022chairs} & ICCV 2023 & Hybrid & SMPL-X & Mesh & 46 & 81 & None \\
        \rowcolor{myLightYellow}\omit & FullBodyManipulation \cite{li2023object} & SIGGRAPH Asia 2023 & Hybrid & SMPL-X & SDF & 17 & 15 & None \\
        & CoChair \cite{liu2023interactive} & arXiv 23.12 & Marker-Based & SMPL-X & CAD & 10 & 8 & None \\
        \rowcolor{myLightYellow}\omit & HOI-M\textsuperscript{3} \cite{zhang2024hoi} & CVPR 2024 & Hybrid & SMPL & Mesh & 31 & 90 & None \\
        & IMHD\textsuperscript{2} \cite{zhao2024m} & CVPR 2024 & Hybrid & SMPL-H & Mesh & 15 & 10 & None \\      
        \rowcolor{myLightYellow}\omit & HIMO \cite{lv2024himo} & ECCV 2024 & Hybrid & SMPL-X & BPS & 34 & 53 & None \\
        & ParaHome \cite{kim2024parahome} & arXiv 24.6 & Hybrid & SMPL-X & Mesh & 30 & 22 & Contact Part \\      
        \rowcolor{myLightYellow}\omit & CORE4D \cite{zhang2024core4d} & arXiv 24.6 & Hybrid & SMPL-X & Mesh & - & 37+ 3K & None \\
        & FORCE \cite{zhang2024force} & 3DV 2025 & Hybrid & SMPL & Mesh & - & 8 & None \\

      \addlinespace[-1.0pt]
      \midrule
      \addlinespace[1.0pt]
      \multirow{19}{*}{\makecell{\rotatebox{90}{\textbf{HHI Datasets}}}}
      & SBU \cite{yun2012two} & CVPR 2012 & Markerless & 3D Skeleton & None & 2 & None & Action Label \\
      \noalign{\global\renewcommand{\arraystretch}{1.3}}

      \rowcolor{myLightYellow}\omit & K3HI \cite{hu2013efficient} & Math. Probl. Eng. 2013 & Markerless & 3D Skeleton & None & 2 & None & Action Label \\      
      & ShakeFive2 \cite{van2016spatio} & ECCV 2016 & Markerless & 3D Skeleton & None & 2 & None & None \\    
      \rowcolor{myLightYellow}\omit & 2C \cite{shen2019interaction} & TVCG 2019 & Synthetic & 3D Skeleton & None & 2 & None & Action Label \\
      & DuetDance \cite{kundu2020cross} & WACV 2020 & Synthetic & 3D Skeleton & None & 2 & None & None \\
      \rowcolor{myLightYellow}\omit & You2me \cite{ng2020you2me} & CVPR 2020 & Markerless & 3D Skeleton & None & 2 & None & None \\
      & CHI3D \cite{fieraru2020three, fieraru2023reconstructing} & CVPR 2020, arXiv 23.8 & Hybrid & 3D Skeleton & None & 2 & None & Text \\
      \rowcolor{myLightYellow}\omit & NTU RGB+D 120 \cite{liu2019ntu} & TPAMI 2020 & Markerless & 3D Skeleton & None & 1-2 & Yes & None \\
      & MultiHuman \cite{zheng2021deepmulticap} & ICCV 2021 & Markerless & SMPL & None & 1-3 & None & None \\
      \rowcolor{myLightYellow}\omit & ExPI \cite{guo2022multi} & CVPR 2022 & Hybrid & 3D Skeleton & None & 2 & None & None \\
      & Hi4D \cite{yin2023hi4d} & CVPR 2023 & Markerless & SMPL & None & 2 & None & None \\
      \rowcolor{myLightYellow}\omit & GTA-Combat \cite{xu2023actformer} & ICCV 2023 & Synthetic & 3D Skeleton & None & 2-5 & None & Action Label \\
      & HHI \cite{liu2023interactive} & arXiv 23.12 & Marker-Based & SMPL-X & Yes & 2 & Yes & Action Label \\
      \rowcolor{myLightYellow}\omit & InterHuman \cite{liang2024intergen} & IJCV 2024 & Markerless & SMPL & None & 2 & None & Text \\
      & DD100 \cite{siyao2024duolando} & ICLR 2024 & Hybrid & SMPL-X & None & 2 & Yes & Music \\
      \rowcolor{myLightYellow}\omit & Inter-X \cite{xu2024inter} & CVPR 2024 & Hybrid & SMPL-X & Yes & 2 & Yes & Text \\
      & ReMoCap \cite{ghosh2024remos} & ECCV 2024 & Markerless & SMPL-X & None & 2 & Yes & None \\
      \rowcolor{myLightYellow}\omit & WebVid-Motion \cite{shan2024towards} & ECCV 2024 & Synthetic & SMPL & None & 2+ & None & Text \\
      & InterDance \cite{li2024interdance} & arXiv 24.12 & Hybrid & SMPL-X & Yes & 2 & Yes & Music \\

      \addlinespace[-2pt]
      \bottomrule[0.4mm]
\label{datasets}
\end{longtable}
\twocolumn

\subsection{Human-Human Interaction Datasets}
Human-human interaction motion generation datasets include interaction motions involving multiple individuals, provided with detailed annotations such as action labels, text descriptions, asymmetry annotations, and other relevant interaction features.

\textbf{SBU} \cite{yun2012two} contains motion capture data of two-person interactions recorded with Kinect. It comprises 300 interactions performed by 7 participants in 21 unique two-person pairings, annotated frame-by-frame with action labels.

\textbf{K3HI} \cite{hu2013efficient} provides 320 two-person interactions performed by 15 participants, categorized into six types of complex actions. The dataset is annotated with 3D skeleton data, capturing 15 joints per person.

\textbf{ShakeFive2} \cite{van2016spatio} includes fine-grained dyadic human interactions recorded with Kinect, covering five interaction types: fist bump, handshake, high five, hug, and passing an object. The dataset includes 3D joint position annotations for both participants, allowing spatio-temporal analysis of body movement coordination.

\textbf{2C} \cite{shen2019interaction} consists of 95 synthetic kickboxing interactions with a duration of 206 seconds, generated by synthesizing shadow-boxing motions into semantic interaction classes based on predefined patterns.

\textbf{DuetDance} \cite{kundu2020cross} contains poses represented as 3D skeletons extracted from YouTube videos of complex duet dances, featuring synchronized complementary movements.

\textbf{You2Me} \cite{ng2020you2me} captures one-on-one interaction recorded using chest-mounted GoPro cameras, with motions represented as 3D joint positions of the camera wearer and interactee obtained from Kinect V2 systems.

\textbf{CHI3D} \cite{fieraru2020three} contains 631 interaction motion sequences, featuring actions like handshake, hug, and push with detailed contact annotations, providing 3D skeletal representations and contact maps for close human interactions.

\textbf{NTU RGB+D 120} \cite{liu2019ntu} includes 114K RGBD videos collected using Kinect v2, featuring 106 participants performing 120 distinct actions under 96 backgrounds and varied camera views. The dataset includes 3D human skeleton motion data with 25 joint coordinates per person.

\textbf{MultiHuman} \cite{zheng2021deepmulticap} provides 150 3D scans of multi-human interaction motions, featuring up to three individuals per scene with varied occlusion levels. Captured using a dense camera rig, the dataset provides textured meshes to enhance multi-human performance under occlusions.

\textbf{ExPI} \cite{guo2022multi} collects 115 sequences of professional dancers performing Lindy Hop aerial steps, recorded by multi-camera motion capture system. It contains 30K frames annotated with 3D body poses and shapes, synchronized multi-view videos, and motion data.

\textbf{Hi4D} \cite{yin2023hi4d} includes 20 subject pairs performing 100 dynamic interaction sequences with over 11K frames. It also provides vertex-level contact annotations, captured with multi-view systems and processed for disentangling interacting subjects.

\textbf{GTA-Combat} \cite{xu2023actformer} consists of 7K synthetic multi-person motion sequences captured from the Grand Theft Auto V gaming engine. It features 2-5 characters performing combat behaviors with high variability, modeled through 3D human skeletons and captured as motion sequences annotated with temporal interactions.

\textbf{HHI} \cite{liu2023interactive} collects 5K full-body motion reaction sequences of human-human interactions, including 30 distinct action categories, performed by 10 pairs of participants. It is designed to allow for clear actor and reactor identification, capturing both body and hand interactions.

\textbf{InterHuman} \cite{liang2024intergen} includes 6K interaction motion sequences with approximately 107M frames of two-person interactions. It also includes 16K natural language annotations.

\textbf{DD100} \cite{siyao2024duolando} contains 117 minutes of duet dance data performed by 5 pairs of professional dancers. The dataset includes 3D motion data in SMPL-X format for both the leader and follower, synchronized with background music.

\textbf{Inter-X} \cite{xu2024inter} includes 11K human interaction sequences across 40 action categories, featuring precise body and hand movements recorded using optical-based motion capture system and inertial gloves. It provides fine-grained textual descriptions and annotations for interaction order, relationships, and personalities.

\textbf{ReMoCap} \cite{ghosh2024remos} provides 275.7K motion frames of two-person interactions, including Lindy Hop dancing and Ninjutsu martial arts, recorded with markerless multi-view motion capture systems. It features full-body and finger-level articulations, 3D skeleton poses, and multiview RGB videos.

\textbf{WebVid-Motion} \cite{shan2024towards} consists of 3.5K video-motion-text tuples from WebVid-10M \cite{bain2021frozen}, featuring multi-person motions annotated in SMPL format, capturing diverse natural poses and interactions in open-domain scenarios.

\textbf{InterDance} \cite{li2024interdance} contains 3.93 hours of duet dance motion data performed by professional dancers across 15 dance genres, including ballroom, folk, classical, and street dance. Captured using an optical motion capture system, the dataset provides SMPL-X motion representations with detailed full-body and finger movements, emphasizing strong interactions such as hand holding and waist holding.

\subsection{Evaluation Metrics}
Evaluation Metrics for 3D human interaction generation can be broadly categorized into three main groups: motion quality, physical plausibility, and user studies. Motion quality metrics focus on evaluating the realism, accuracy, and diversity of generated interaction motions, typically including measures such as FID, Diversity, Multimodality, R Precision, MM Dist, MPJPE, and APD. Physical plausibility metrics evaluate how well generated interactions adhere to physical constraints with environments or objects, commonly measured by Foot Sliding, Non-Collision, Contact, and Penetration. User studies involve subjective assessments from human evaluators to measure the perceived naturalness and overall perceptual quality of generated interactions.

\textbf{FID.} Fréchet Inception Distance measures the distance between the distributions of real and generated motions \cite{guo2020action2motion}. It computes the Fréchet distance between the activations of a pre-trained Inception network on real and generated samples. Lower FID values indicate that the generated samples are closer to real samples in terms of their distribution.

\textbf{Diversity.} Diversity evaluates the variety of the generated human interactions \cite{guo2020action2motion}. It measures how different the generated motions are from each other, ensuring that the generative model produces a wide range of possible outcomes rather than repetitive or similar ones. High diversity indicates a better performance in generating a rich set of distinct motions.

\textbf{Multimodality.} Multimodality evaluates whether the model can produce different types of interaction motions for the same interactive entity, capturing the inherent variability in human behavior \cite{guo2020action2motion}. High modality indicates that the model can produce diverse outcomes across multiple distinct modes.

\textbf{R Precision.} R Precision evaluates the proportion of relevant interaction motions included in the top R generated results \cite{Guo_2022_CVPR}. It measures the accuracy of generation by comparing the number of relevant motions within the first R results. This metric provides an intuitive assessment of how well the model produces relevant outcomes in the top-ranked results.

\textbf{MM Dist.} Maximum Mean Discrepancy Distance measures the distance between the distributions of real and generated interaction motions by comparing their means in a reproducing kernel Hilbert space \cite{Guo_2022_CVPR}. It computes the maximum possible difference between the two distributions, with lower values indicating that the generated samples are closer to the real samples in terms of distribution. 

\textbf{MPJPE.} Mean Per Joint Position Error evaluates the average Euclidean distance between corresponding joints in the predicted and real 3D motions \cite{ionescu2013human3}. It measures the accuracy of motion generation by computing the deviation of each joint position from the ground truth across all frames. A lower MPJPE value indicates a more realistic and precise interaction motion.

\textbf{APD.} Average Pairwise Distance evaluates the diversity of generated results by computing the average Euclidean distance between pairs of generated motion sequences \cite{yuan2020dlow}. A higher APD value refers to greater diversity in the generated interaction motions, which is essential for capturing a wide range of realistic human movements.

\textbf{Foot Sliding} Foot Sliding assesses the realism of generated motions by measuring unintended foot movements when they are expected to remain stationary \cite{he2022nemf}. It measures how well the model maintains foot-ground contact consistency, reducing artifacts where the feet appear to slide unnaturally.

\textbf{Non-Collision.} Non-collision score evaluates the physical plausibility of generated interaction motions by measuring the proportion of body mesh vertices that do not penetrate the environment or other body parts \cite{zhang2020generating}. It is computed as the ratio of body mesh vertices with positive Signed Distance Field values to the total number of vertices. A higher non-collision score indicates more realistic and physically valid motion generation, minimizing self-intersections and environmental collisions.

\textbf{Contact.} Contact score measures the proportion of generated body meshes that maintain expected physical contact with the environment or other objects \cite{zhang2020generating}. It is calculated as the number of body meshes exhibiting contact divided by the total number of generated body meshes. A higher contact score indicates more realistic interaction with the surrounding interactive entities.

\textbf{Penetration.} Penetration score measures the degree of intersection between body meshes and the environment or themselves \cite{zhao2023synthesizing}. It evaluates the extent of unrealistic penetrations, where body parts or external objects overlap unnaturally. A lower penetration score indicates more physically plausible human interaction generation, improving adherence to environmental constraints.

\textbf{User Study.} User study assesses the perceptual quality of generated results based on human evaluation. Participants rate the realism, naturalness, and physical plausibility of interaction sequences and express their preferences through comparative judgments. It provides qualitative insights that complement quantitative metrics, helping to refine motion generation models.

\section{Conclusion and Future Work} \label{Conclusion and Future Work}
In this survey, we provide a thorough overview of recent advancements in human interaction generation. The categorization is based on the interactive entities: human-scene interaction, human-object interaction, and human-human interaction. We also provide an overview of commonly used datasets and evaluation metrics in this area. This survey serves as a comprehensive reference to the advances and challenges in the field, providing valuable insights for future research and practical applications. While the current advancements have brought us closer to understanding human interaction generation, there are still unresolved issues that hinder further progress. To tackle these, we identify a number of prospective research directions that have the potential to advance the field forward.

\textbf{Large-Scale Real-World Data.} Collecting real-world interaction data in diverse scenarios is crucial for enhancing the robustness and adaptability of human interaction generation \cite{zhang2024force, petrov2024tridi, zhang2024hoi}. Unlike readily available 2D data, capturing high-quality 3D data is significantly more resource-intensive and often occurs under uncontrolled conditions—especially when both the human subject and the interactive entities must be recorded simultaneously. Furthermore, effective motion representation techniques are critical to handle the complexity and variability of real-world data. These challenges underscore the importance of advanced motion capture technologies and representation methods to achieve the efficiency and accuracy required for large-scale, real-world datasets.

\textbf{Controllable generation.} Future research could move beyond strongly correlated inputs, such as motion history or text descriptions, by incorporating a broader range of contextual conditions\cite{xu2024inter, li2024zerohsi, li2025controllable}, including human personality, emotional states, and stylistic preferences. Effectively aligning these diverse modalities has the potential to produce motions that capture a richer spectrum of human experiences and behaviors. However, developing model architectures capable of handling diverse input types present significant technical challenges that require further exploration.

\textbf{Multi-Person and Multi-Object Interactions.} Generating realistic interaction motion in crowded or collaborative environments involves simultaneously modeling multiple human \cite{zhang2024hoi, liu2023interactive, zhang2023neuraldome} and multiple objects \cite{he2024syncdiff, liu2023interactive, lv2024himo, xu2023interdiff}. These scenarios often involve group tasks that require precise temporal and spatial coordination, as well as complex interactions engaging various body parts. To address these challenges, models must effectively capture diverse relational dynamics, including joint attention, collision avoidance, and role-specific behaviors, to ensure coherent and natural interactions.

\textbf{Non-rigid Object Interactions.} Beyond rigid-body assumptions, real-world scenarios often involve objects with flexible, articulated, or deformable properties, such as cloth, ropes, soft tissues, and foldable furniture. While articulated objects have gained some attention \cite{he2024syncdiff, xu2021d3d, kim2024parahome, jiang2022chairs}, the modeling and simulation of elastic or deformable objects \cite{song2024hoianimator, cao2024avatargo, guzov2024interaction} remain relatively underexplored. Generating realistic interactions with such objects requires advanced physics engines and sophisticated model representations that can accurately account for shape deformations and material properties.

\textbf{Face and Finger Motion Generation.} High-fidelity human motion includes not only coarse body movements but also subtle facial expressions and intricate finger motions that convey nuanced interactions. Realistically generating these fine-grained details \cite{jiang2024autonomous, lv2024himo, yan2023cimi4d, yin2023hi4d} requires high-resolution data capture and advanced modeling techniques capable of learning and reproducing subtle motion features. Incorporating such detailed expressions and hand poses is particularly critical for close-up interactions, as they significantly enhance the realism and accuracy of the generated motions.

\textbf{Human-Scene-Object-Human Interactions.} Beyond single-type interactive entities, motion generation in extensive environments \cite{zhang2024hoi, li2024zerohsi, liu2024revisit, jiang2024scaling, huang2022capturing} requires managing dynamic surroundings, including objects, scenes, and other humans. A unified model capable of flexibly updating environmental changes in response to human motion while maintaining physical plausibility is critical. Scaling up human interaction generation in open and complex environments will rely on effectively integrating scene understanding, path planning, object manipulation, and multi-human interaction within a cohesive framework.

\ifCLASSOPTIONcaptionsoff
  \newpage
\fi



%
\vspace{-4pt}
{\small
\bibliographystyle{IEEEtran}

\bibliography{references}
}

\end{document}